\documentclass[a4paper,10pt]{article}

\usepackage[top=1.5in, bottom=1.5in, left=1in, right=1in]{geometry}
\usepackage[dvips]{graphicx}
\usepackage{url}

\usepackage{amssymb,amsfonts,amsmath}
\newcommand{\bx}{\mathbf{x}}

\begin{document}

\title{A theoretical basis for efficient computations 
\\with noisy spiking neurons}
\author{\small Zeno Jonke$^\dagger$, Stefan Habenschuss$^\dagger$, Wolfgang Maass\footnote{Corresponding author: maass@igi.tugraz.at. $^\dagger$ These authors contributed equally to this work. }\\\small 
Institute for Theoretical Computer Science, Graz University of Technology, Austria}

\maketitle

{\small
\textbf{Abstract.} 
Network of neurons in the brain apply -- unlike processors in our current generation of computer hardware -- an event-based processing strategy, where short pulses (spikes) are emitted sparsely by neurons to signal the occurrence of an event at a particular point in time. Such spike-based computations promise to be substantially more power-efficient than traditional clocked processing schemes. However it turned out to be surprisingly difficult to design networks of spiking neurons that are able to carry out demanding computations. We present here a new theoretical framework for organizing computations of networks of spiking neurons. In particular, we show that a suitable design enables them to solve hard constraint satisfaction problems from the domains of planning/optimization and verification/logical inference. The underlying design principles employ noise as a computational resource. Nevertheless the timing of spikes (rather than just spike rates) plays an essential role in the resulting computations. Furthermore, one can demonstrate for the Traveling Salesman Problem a surprising computational advantage of networks of spiking neurons compared with traditional artificial neural networks and Gibbs sampling. The identification of such advantage has been a well-known open problem.}

\vspace{1cm}

The number of neurons in the brain lies in the same range as the number of transistor in a supercomputer. But whereas the brain consumes less then 30 Watt, a supercomputer consumes as much energy as a major part of a city. The power consumption has not only become a bottleneck for supercomputers, but for many applications and improvements of computing hardware, including the design of intelligent mobile devices.  But how can one capture the drastically different style of computations by networks of neurons in the brain, and apply similar energy-efficient methods for the organization of computation in novel computing hardware? In particular, how can one carry out complex computations in massively parallel systems without a clock, that synchronizes the contributions of individual processors?

When the membrane potential of a biological neuron crosses a threshold, the neuron emits a spike, i. e. a sudden voltage increase that lasts for 1 - 2 ms. Spikes occur asynchronously in continuous time and are communicated to numerous other neurons via synaptic connections with different strengths (``weights''). The effect of a spike from a pre-synaptic neuron $l$ on a post-synaptic neuron $k$, the so-called post-synaptic potential (PSP), can be approximated as an additive contribution to its membrane potential. It is short-lived (10 - 20 ms) and can be either inhibitory or excitatory, depending on the sign of the synaptic weight $w_{kl}$. 
It is a long-standing mystery how complex computations are organized in networks of spiking neurons in the brain, especially in view of ubiquitous sources of noise in neurons and synapses, and large  trial-to-trial variability of network responses~\cite{FaisalETAL:08}. 
Hence it is not surprising, that attempts to port brain-inspired computational architectures into novel artificial computing hardware (see e.g. \cite{Mead:90,Cauwenberghs:13,PfeilETAL:13,HamiltonETAL:14,McDonnellETAL:14}) have had only limited success from the computational perspective. Whereas very nice results were achieved with artificial spike-based retinas \cite{LichtsteinerETAL:08} and cochleas \cite{ChanETAL:07}, we are not aware of published methods for solving complex computational tasks efficiently by spike-based circuits. 
Powerful computations with spiking neurons have previously been demonstrated in \cite{EliasmithETAL:12}. But spiking neurons were used there in a rate-coding mode, and could be replaced by standard non-spiking artificial neural network units. In this article we want to address the additional challenge to find computational uses of spiking neurons where the relative timing of single spikes is used, rather than their firing frequency.

We will focus on computations with noisy spiking neurons as a contribution to the emergent new field of stochastic electronics \cite{HamiltonETAL:14}. A hardware emulation of these methods will therefore have to make use of efficient methods for generating random numbers in hardware, which are not addressed here (see e.g. \cite{TetzlaffETAL:12,YangETAL:14} for some new approaches). Finally we would like to clarify that this article does not aim at modelling spike-based computations in biological organisms. We also would like to emphasize, that there are many well-known methods for solving the computational tasks considered in this article very efficiently in software and hardware. The methods that we are introducing are only of interest under the premise that one wants to employ spike-based hardware for computations, e.g. because of its power-efficiency \cite{Mead:90,MerollaETAL:14}.

We propose in this article a new theoretical basis and principles for the design of computationally powerful spike-based networks. Rather than attempting to design such networks on the basis of presumed ``neural codes'' for salient variables, and computational operations together with signal processing schemes for such neural codes, we propose to focus instead on the statistics and dynamics of network states.
These network states record which neurons fire within some small time window.
\begin{figure}[tbp]
  \begin{center}
\includegraphics[width=\textwidth]{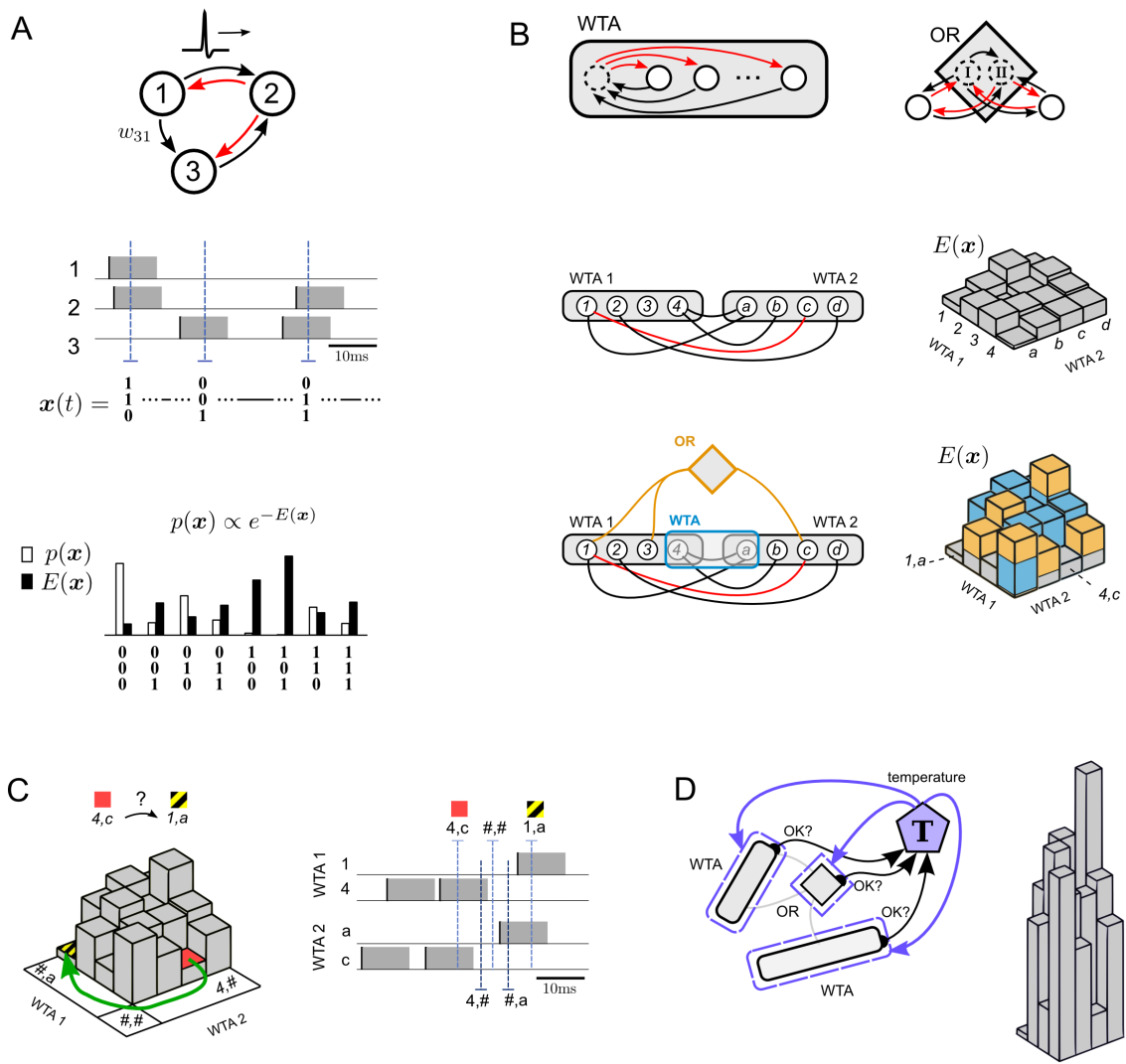}
\caption{Illustration of four principles for solving hard constraint satisfaction problems by spike-based neural networks with noise. \textbf{A}. Focus on the probability $p(\bx)$ (or energy $E(\bx)$) of network states $\bx$, where each spike of a neuron switches a corresponding component of $\bx$ to 1 for a time window of length $\tau$ that corresponds to the duration of a PSP (black/red arrows=excitatory/inhibitory connections).  \textbf{B}. Modularity principle for the energy function $E(\bx)$. The energy landscape of a subnetwork of principal neurons (white circles) can be shaped by adding circuit motifs (such as WTA, OR) with auxiliary neurons (dashed circles) that provide approximately linear contributions to $E(\bx)$. In addition, Winner-Take-All (WTA) circuits allow to emulate arbitrary multi-nomial problem variables, see WTA1 (for a variable ranging over $\{1,2,3,4\}$) and WTA2 (for a variable ranging over $\{a,b,c,d\}$) in the bottom plot. Lines without arrowheads denote symmetric connections. \textbf{C}. Spike-based dynamics supports bypassing of high-energy barriers. This example illustrates bypassing of energy barriers between low energy states $4,c$ (red) and $1,a$ (yellow/black) for two problem variables by moving through network states that represent undefined values of these problem variables. Note that these intermediate network states have actually high energy, but are nevertheless frequently visited as shown in section \ref{sec:2.4}. \textbf{D}. Internal spike-based temperature control can drastically change the contrast of the energy landscape from C on the basis of internal information about the satisfaction of salient constraints, and thereby the progress of stochastic search.}
\label{fig:1}
  \end{center}
\end{figure}
In this framework the fine-scale dynamics of recurrently connected networks of spiking neurons with noise can be interpreted as stochastic search for low energy network states (principle 1, Fig.~\ref{fig:1}A). This energy landscape can be shaped for concrete computational tasks in a modular fashion by composing the network from simple stereotypical network motifs (principle 2, Fig.~\ref{fig:1}B). Furthermore, we show that spike-based signaling allows to bypass barriers in the energy landscape, thereby elucidating specific advantages of spike-based computation (principle 3, Fig.~\ref{fig:1}C). In particular, we provide a theoretical basis for understanding why spike-based search for low energy network states requires in some cases fewer steps than in Boltzmann machines. Finally, we exploit that networks of spiking neurons can internally rescale their energy landscape in order to lock into desirable solutions (principle 4, Fig.~\ref{fig:1}D).
Altogether these design principles and the underlying computational theory suggest a stark departure from previous approaches for designing computationally powerful networks of spiking neurons. In particular, they suggest new methods for solving constraint satisfaction problems in energy efficient spike-based electronic systems.

\section*{Results}

As usual, we model the stochastic behavior of a spiking neuron $k$ at time $t$ via an instantaneous firing probability (i.e.~probability of emitting a spike), 
\begin{eqnarray}
\frac{1}{\tau}\; \exp({u_k(t)})\;\;,\label{eq:neuronmodelrho}
\end{eqnarray}
that depends on the current \emph{membrane potential} $u_k(t)$ of the neuron. It is defined as the weighted sum of the neuron's inputs,
\begin{eqnarray}
u_k(t) = b_k + \sum_{l} w_{kl}\,x_l(t)\;\;.\label{eq:uk}
\end{eqnarray}
The additional bias term $b_k$ represents the intrinsic excitability of neuron $k$. $w_{kl} x_l(t)$ models the PSP at time $t$ that resulted from a firing of the pre-synaptic neuron $l$. 
This is a standard model for capturing stochastic firing of biological neurons \cite{JolivetETAL:06}.
We assume here for mathematical tractability rectangular PSP shapes $x_l(t)$ of length $\tau$ ($\tau=10$ms throughout this paper), i.e. $x_l(t)=1$ if a spike was emitted by neuron $l$ within $(t-\tau, t]$, and $x_l(t)=0$ otherwise~(Fig.~\ref{fig:1}A). We say then that neuron $l$ is in the active state (or ``on'' state) during this time interval of length $\tau$ (which coincides with its refractory period). For simplicity we assume that the refractory period during which a neuron cannot spike equals~$\tau$, the length of a PSP.

\subsection*{Network states, stationary distributions and energy functions}

We propose to focus on the temporal evolution and statistics of spike-based network states (principle 1), rather than on spikes of individual neurons, or rates, or population averages. The \emph{network state} of a network of $N$ neurons at time~$t$ is defined like in \cite{BerkesETAL:11} as $\mathbf{x}(t)=(x_1(t), x_2(t), \dots, x_N(t))$ (Fig.~\ref{fig:1}A, middle), where $x_k (t) = 1$ indicates that neuron $k$ has fired (i.e., emitted a spike) during the time interval $(t - \tau, t]$ corresponding to the duration of a PSP. Else $x_k(t) = 0$. 
If there is a sufficient amount of noise in the network (as e.g. injected through the stochastic firing rule in equation \eqref{eq:neuronmodelrho}), the distribution of these continuously varying network states converges exponentially fast from any initial state to a unique stationary (equilibrium) distribution $p(\bx)$ of network states \cite{HabenschussETAL:13}. $p(\bx)$ can be viewed as a concise representation of the statistical fine-structure of network activity at equilibrium. In line with related non-spiking network analyzes \cite{Hopfield:86} we will use in the following an alternative representation of $p(\bx)$, namely the energy function $E (\bx) = - \log p(\bx) + C$, where $C$ denotes an arbitrary constant (Fig.~\ref{fig:1}A bottom), so that low energy states occur with high probability at equilibrium.

The \emph{neural sampling} theory \cite{BuesingETAL:11} implies that the stationary distribution of a network with neuron model given by equations \eqref{eq:neuronmodelrho} and \eqref{eq:uk}, and symmetric weights $w_{kl}=w_{lk}$ is a Boltzmann distribution with energy function
\begin{eqnarray}
E_{\mathcal{N}}(\mathbf{x}) = - \sum_{k=1}^N b_k x_k - \frac{1}{2} \sum_{k=1}^N \sum_{l=1}^N x_k x_l w_{kl}\;\;.\label{eq:boltze}
\end{eqnarray}
This energy function has at most second-order terms. Many constraint satisfaction problems (such as 3-SAT, see below), however, require the modeling of higher-order dependencies among problem variables. To introduce higher-order dependencies among a given set of \emph{principal} neurons $\mathbf{x}$, one needs to introduce additional \emph{auxiliary} neurons $\boldsymbol{\xi}$ to emulate the desired higher-order terms. Two basic approaches can be considered. In the first approach, the connections between the principal neurons $\mathbf{x}$ and the auxiliary neurons $\boldsymbol{\xi}$, as well as the connections within each group, are constrained to be symmetric. In such case, the network as a whole is symmetric and hence the energy function $E_{\mathcal{N}}(\mathbf{x},\boldsymbol{\xi})$ of the joint distribution over principal neurons and auxiliary neurons can be described with at most second order terms. The marginal energy function of the principal neurons,
\begin{align}
E_{\mathcal{N}}(\mathbf{x}) = \log \sum_{\boldsymbol{\xi}} \exp( E_{\mathcal{N}}(\mathbf{x},\boldsymbol{\xi}))
\end{align}
will generally feature complex higher-order terms.
By clever use of symmetrically connected auxiliary neurons one may thereby introduce arbitrary higher-order dependencies among principal neurons. In practice, however, this ``symmetric'' approach has been found to prohibitively slow down convergence to the stationary distribution \cite{Pecevski2011}, due to large energy barriers introduced in the energy landscape when one introduces auxiliary variables through deterministic definitions.

The alternative approach, which is pursued in the following, is to maintain symmetric connections among principal network neurons, but to abandon the constraint on symmetricity for connections between principal and auxiliary neurons, as well as for connections among auxiliary neurons. 
Furthermore auxiliary variables or neurons are related by stochastic (rather than deterministic) relationships to principal neurons.
The theoretical basis for constructing appropriate auxiliary network motifs is provided by the neural computability condition (NCC) of \cite{BuesingETAL:11}. The NCC states that it suffices for a neural emulation of an arbitrary probability distribution $p(\bx)$ over binary vectors $\bx$ that there exists for each binary component $x_k$ of $\bx$ some neuron $k$ with membrane potential 
\begin{align}
 u_k(t) = \log \frac{p(x_k = 1|\mathbf{x}_{\setminus k} (t))}{p(x_k = 0|\mathbf{x}_{\setminus k}(t))}\;\;,\label{eq:NCC}
\end{align}

\noindent where $\bx_{\setminus k}(t)$ denotes the state of all neurons except neuron $k$. 
For a second-order Boltzmann distribution, evaluating the right-hand side gives the simple linear membrane potential in equation~\eqref{eq:uk}. For more complex distributions, additional higher-order terms appear. 

\subsection*{Modularity of energy function}

The shaping of the energy function of a network of spiking neurons can be drastically simplified through a modularity principle (principle 2). It allows us to understand the energy function $E(\bx)$ of a large class of networks of spiking neurons in terms of underlying generic network motifs. 

As introduced above, we distinguish between principal neurons and auxiliary neurons: Principal neurons constitute the interface between network and computational task. For example, principal neurons can directly represent the variables of a computational problem, such as the truth values in a logical inference problem (Fig.~\ref{fig:4}). The state $\bx(t)$ of the \emph{principal network} (i.e., the principal neurons) reflects at any moment $t$ an assignment of values to the problem variables. Auxiliary neurons, on the other hand, appear in specialized network motifs that modulate the energy function of the principal network. More specifically, the purpose of auxiliary neurons is to implement higher-order dependencies among problem variables. The starting point for constructing appropriate auxiliary circuits is the NCC (equation \eqref{eq:NCC}) rewritten in terms of energies,

\begin{eqnarray}
u_k (t) = E \left(x_k=0, \bx_{\setminus k}(t)\right) - E \left(x_k = 1, \bx_{\setminus k}(t)\right)\;\;.\label{eq:mainuk}
\end{eqnarray}
\noindent

\noindent This sufficient condition allows to engage additional auxiliary neurons, which are not subject to the constraint given by equation \eqref{eq:mainuk}, in order to shape the energy function $E (\bx)$ of the principal network in desirable ways:
Suppose that a set of auxiliary circuits $\mathcal{I}$ is added (and connected with symmetric or asymmetric connections) to a principal network with energy function given by equation \eqref{eq:boltze}. Due to linearity of membrane integration ( equation~\eqref{eq:uk}), the membrane potential of a principal neuron $k$ in the presence of such auxiliary circuits can be written as,
\begin{align}
u_{k}(t) = b_k + \sum_{l=1}^N w_{kl}\,x_l(t) + \sum_{i \in  \mathcal{I}} u_{k,i}(t) \;\;,\label{eq:ukaux}
\end{align}
where the instantaneous impact of auxiliary circuit $C_i$ on the membrane potential of principal neuron $k$ is denoted by $u_{k,i}(t)$. In the presence of such auxiliary circuits there is no known way to predict the resulting stationary distribution $p(\bx)$ over principal network states (or equivalently $E(\bx)$) \emph{in general}. Condition given by equation \eqref{eq:mainuk}, however, implies that if certain design rules are observed, each auxiliary motif makes a predictable \emph{linear} contribution to the energy function $E (\bx)$ of the principal network (see blue and yellow circuit motif in Fig.~\ref{fig:1}B bottom). \\

\textbf{Theorem 1 (Modularity Principle).} \emph{Let $\mathcal{N}$ be a network of stochastic neurons $k=1,\dots, N$ according to equations \eqref{eq:neuronmodelrho} and \eqref{eq:uk}, symmetric connections $w_{kl}=w_{lk}$ (but no self-connections, i.e.~$w_{kk}=0$) and biases $b_k$. In the absence of auxiliary circuits this principal network has an energy function $E_{\mathcal{N}}(\mathbf{x})$ with first- and second-order terms as defined in equation \eqref{eq:boltze}. Let \mbox{$\mathcal{C} = \{C_1, \dots, C_L\}$} be a set of $L$ additional auxiliary circuits which can be reciprocally connected to the principal network $\mathcal{N}$ to modulate the behavior of its neurons.  Suppose that for each auxiliary circuit $C_i$ there exists a function $U_i(\mathbf{x})$ such that at any time $t$ the following relation holds,
\begin{align}
     u_{k,i}(t) &= U_i\left(x_k=0, \mathbf{x}_{\setminus k}(t)\right) - U_i\left(x_k=1, \mathbf{x}_{\setminus k}(t)\right)\;\;,\label{eq:deltaudeltae}
\end{align}
for any neuron $k$ in $\mathcal{N}$. The relation in equation \eqref{eq:deltaudeltae} is assumed to hold for each auxiliary circuit $C_i$ regardless of the presence or absence of other auxiliary circuits. Then the modulated energy function $E_{\mathcal{N}, \mathcal{I}}(\mathbf{x})$  of the network in the presence of some arbitrary subset $\mathcal{I} \subseteq \{1, \dots, L\}$ of auxiliary circuits can be written as a linear combination,
\begin{align}
 E_{\mathcal{N}, \mathcal{I}}(\mathbf{x}) = E_{\mathcal{N}}(\mathbf{x}) + \sum_{i \in  \mathcal{I}} U_i(\mathbf{x})~~~.
\end{align}
}\\

Examples for network motifs that impose computationally useful higher order constraints in such modular fashion are the winner-take-all (WTA) and OR motif (Fig.~\ref{fig:1}B). 
The WTA motif (see Supplementary material section 3.2) is closely related to ubiquitous motifs of cortical microcircuits \cite{DouglasMartin:04}. It increases the energy of all network states where not exactly one of the $K$ principal neurons, to which it is applied, is active (this can be realized through an auxiliary neuron mediating lateral inhibition among principal neurons).
It may be used, for example, to enforce that these $K$ principal neurons represent a discrete problem variable with $K$ possible values. The two WTAs shown in Fig.~\ref{fig:1}B (middle left), for example, represent two discrete variables with values $1,2,3,4$ and $a,b,c,d$. The energy of any combination of their values is shown on the right. The red (inhibitory) connection between $1$ and $c$, for example, causes the highest energy for the joint state $(1,c)$.
The OR-motif, which can be applied to any set of principal neurons, enforces that most of the time at least one of these principal cells is active. It can be implemented through two auxiliary neurons $I$ and $II$, which are reciprocally connected to these principal neurons (as illustrated in Fig.~\ref{fig:1}B for two principal neurons). Neuron $I$ excites them, and neuron $II$ curtails this effect through subsequent inhibition as soon as one of them fires (see Supplementary material section 3.3)).

\begin{figure}[tbp]
  \begin{center}
  \includegraphics[width=\textwidth]{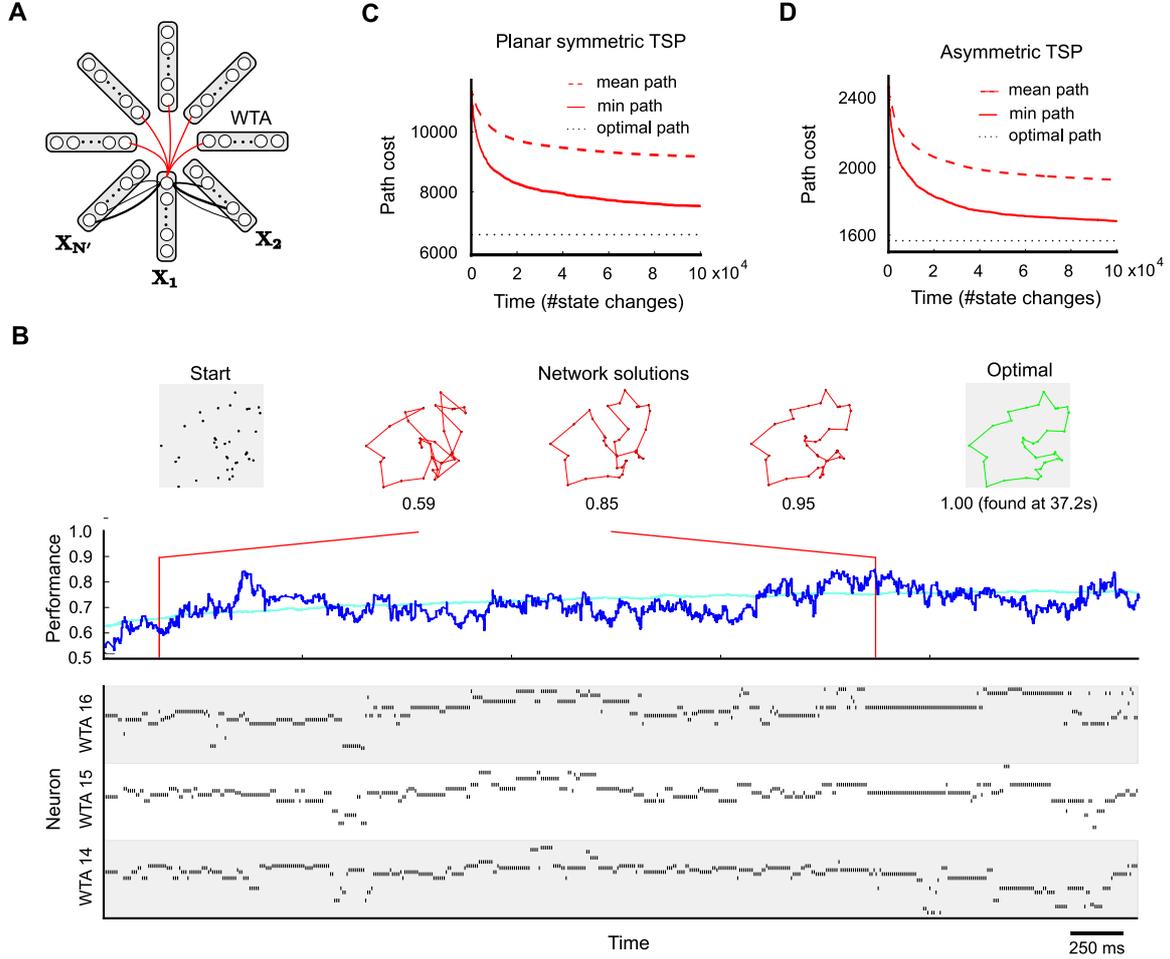}
\caption{Application to TSP. \textbf{A}. Network design, with one WTA circuit $X_n$ for each of the \mbox{$N'=N+N_{resting}$} steps of a tour through $N$ cities (see main text). The synaptic weights between two steps are chosen to reflect movement costs between each pair of cities. The constraint that each city must be visited exactly once is enforced by inhibitory connections among neurons coding for the same city at different time steps. \textbf{B}. Example application to a planar TSP instance with $N=38$. Top: 38-city problem; tours generated by the network; optimal solution. Bottom: spike trains of neurons in selected WTA circuits during the first few seconds of a typical run. Middle: network performance over time (dark blue: single trial, cyan: average over 100 trials). The network quickly generates good approximate solutions to the TSP problem. \textbf{C}. Mean and minimum path length (cost) of solutions generated within a given number of state changes (average over 100 runs; a state change happens whenever a neuron spikes or its PSP ends $\tau$ time units later), for the planar 38-city problem of B. \textbf{D}. Performance for asymmetric TSP problem ($N=39$).}
\label{fig:2}
\end{center}
\end{figure}

\subsection*{Application to the Traveling Salesman Problem}

We demonstrate the computational capabilities that spiking networks gain through these principles in an application to the Traveling Salesman Problem, a well-known difficult (in fact NP-hard) benchmark optimization problem. TSP is a planning problem where given analog parameters $c_{ij}$ represent the cost to move directly from a node (``city'') $i$ in a graph to node $j$. The goal is to find a tour that visits all $N$ nodes in the graph exactly once, with minimal total cost. 

We translate a TSP problem with $N$ nodes (cities) into a network of spiking neurons by representing each step of a tour by a WTA circuit with $N$ neurons (Fig.~\ref{fig:2}A). A few ($N_{resting}$) extra steps (i.e.~WTA circuits) are introduced to create an energy landscape with additional paths to low energy states. Cases where the costs are symmetric ($c_{ij} = c_{ji}$), or even represent Euclidean distances between nodes in 2D, are easier to visualize (Fig.~\ref{fig:2}B, top) but also computationally easier. Nevertheless, also general TSP problems with asymmetric costs can be solved approximately by spike-based circuits (Fig.~\ref{fig:2}D).

\subsection*{Advantage of spike-based stochastic search}\label{sec:2.4}

While the design of network motifs benefits already from the freedom to make synaptic connections in spike-based networks asymmetric (consider e.g. in- and outgoing connections of the auxiliary WTA neuron that implements lateral inhibition), our third principle proposes to exploit an additional generic asymmetry of spike-based computation. A spike, which changes the state of a neuron $k$ to its active state, occurs randomly according to equation~\eqref{eq:neuronmodelrho}. But its transition back to the inactive state occurs (deterministically) $\tau$ time units later. As a result, it may occur for brief moments that all $K$ principal neurons of a WTA motif are inactive, rendering an associated $K$-valued problem variable to be intermittently undefined. Most of the time this has no lasting effect because the previous state is quickly restored. 
But when the transition to an undefined variable state occurs in several WTA circuits at approximately the same time, the network state can bypass high-energy barriers and explore radically different state configurations (Fig.~\ref{fig:1}C). Our theoretical analysis implies that this effect enhances exploration in spike-based networks, compared with standard sampling approaches (Boltzmann machines, Gibbs sampling). The TSP is a suitable study case for such comparison, because we can compare the dynamics of a spiking network with  that of a Boltzmann machine which has exactly the same stationary distribution (i.e. energy function).

\subsubsection*{Specific properties of spike-based stochastic dynamics}

Consider a Boltzmann machine or Gibbs sampler \cite{brooks2010handbook} (operating in continuous time to allow for
a fair comparison; for details see Supplementary material section 4.4) that samples from the same distribution $p(\bx)$ as a given
spiking network. Such non-spiking Gibbs sampler has a symmetric transition dynamics: it activates
units proportional to the sigmoid function $\sigma (u) = (1 + exp(−u))^{−1}$, while off-transitions occur at a rate
proportional to $\sigma (−u)$. Neural sampling in a stochastic spiking network, on the other hand, triggers on-transitions
proportional to $exp (u)$, while off-transitions occur at a constant ``rate''. In neural sampling,
the mean transition times $m_{on}$ from the last \emph{on}$\rightarrow$\emph{off} to the next \emph{off}$\rightarrow$\emph{on} transition and its dual $m_{off}$ in
a neuron with membrane potential $u$ are given by:

\begin{align}
 m_{\textnormal{on}}(u) &= \tau\cdot \exp(-u)\;\;,\\
m_{\textnormal{off}}(u)&=\tau\;\;.
\end{align}

\begin{figure}[!tbp]
  \begin{center}
  \includegraphics[width=0.8\textwidth]{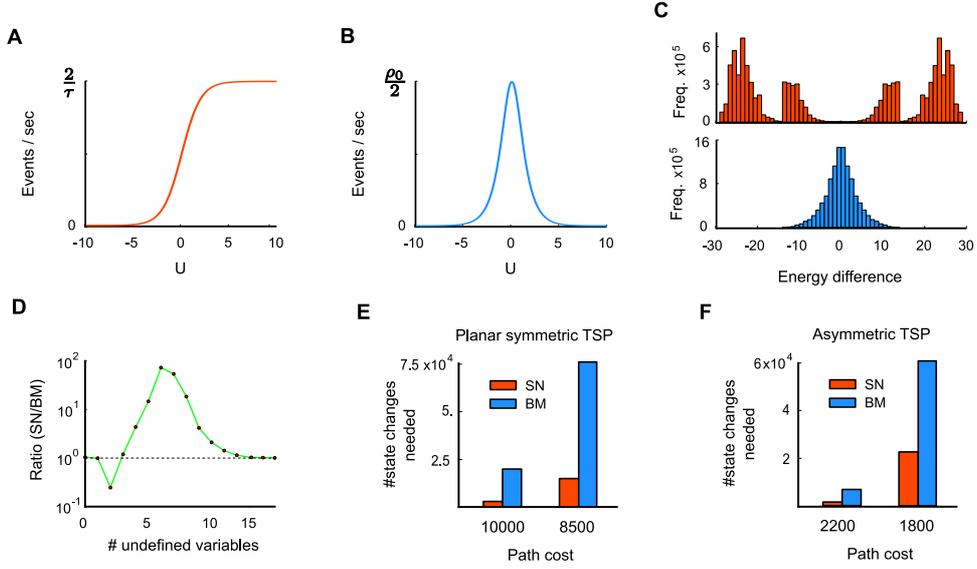}
\caption{Advantage of spike-based stochastic search (third principle). \textbf{A}. The event rate (counting both on and off events) of a spiking neuron is low for small membrane potentials $u$, and saturates for high $u$. \textbf{B}. In contrast, the event rate of a Boltzmann unit is sharply peaked around $u=0$. \textbf{C}. Histogram of energy jumps due to state changes, computed from all neural on- and off-transitions in the network (top red: SN, bottom blue: BM). \textbf{D}. Comparison between SN and BM (SN=spiking network, BM=Boltzmann machine) regarding the frequency of state transitions that leave one or several problem variables undefined. Shown is the ratio SN/BM for the symmetric TSP problem considered in Fig.~\ref{fig:2}. Transitions into states with several undefined variables occur with higher frequency in the SN compared to the BM. \textbf{E}. Comparison of typical computation times (in terms of the number of state changes of the network) until a tour with a given cost is found for the problem instance from Fig.~\ref{fig:2}B, for spike-based networks and corresponding Boltzmann machines. \textbf{F}. Same for asymmetric 39 city problem considered in Fig.~\ref{fig:2}D. The results indicate specific advantages of spike-based stochastic search.}
\label{fig:3}
 \end{center}
\end{figure}

On average, an off-and-on (or on-and-off) transition sequence takes $m_{on} (u) + m_{off} (u)$ time units. Thus,
the average event rate $R(u)$ at which a spiking neuron with membrane potential $u$ changes its state is
given by (Fig.~\ref{fig:3}A),
\begin{align}
R(u)=\frac{2}{m_{on}(u) + m_{off}(u)} = \frac{2}{\tau} \sigma(u)~~~~~.
\end{align}

The average event rate $R^{sym}(u)$ in a Gibbs sampler at membrane potential $u$ is given by (Fig.~\ref{fig:3}B),

\begin{align}
R^{sym}(u) = \frac{2 \rho_0}{2 + exp(u) + exp (-u)}~~~~~,
\end{align}

\noindent where $\rho_0$ is a positive constant which controls the overall speed of sampling in the continuous-time Gibbs sampler, with the default choice being $\rho_0=1$.
Clearly, although asymmetric spiked-based and symmetric Gibbs sampling both sample from the same distribution $p(\bx)$ of states of the principal network, the frequency of state transitions at different levels of the membrane potential $u$ differs quite drastically
between the two samplers. Concretely, the following u-dependent factor $F(u)$ relates the two systems:
\begin{align}
F(u) = \frac{R(u)}{R^{sym}(u)} = \frac{1}{\underbrace{\tau \rho_0}_{\text{const.}}} \cdot \left( 1 + \exp(u) \right)\;\;.
\end{align}

\noindent Similar to $\tau$ in the spike-based system, $\rho_0$ can be understood as a global time scale parameter which has no bearing on the fine-scale dynamics of the system. The remaining factor reveals a specific $u$-dependence of transition times which greatly affects the fine-scale dynamics of sampling. 
Note that $F(u)$ is strictly positive and increases monotonically with increasing membrane potential~$u$.
Hence, the asymmetric dynamics of spiking neurons increases specifically the \emph{on}- and \emph{off}-transition rates of neurons with high membrane potentials~$u$ (i.e.~neurons with strong input and/or high biases). According to equation \eqref{eq:mainuk}, however, high membrane potentials $u$ reflect large energy barriers in the energy landscape. Therefore, the increase of transition rates for large $u$ in the spike-based system (due to the asymmetry introduced by deterministic \emph{on}$\rightarrow$\emph{off} transitions) means that large energy barriers are expected to be crossed more frequently than in the symmetric system (see Supplementary material section 4 for further details).

\subsubsection*{Spikes support bypassing of high energy barriers}

A specific computational consequence of spike-based stochastic dynamics is demonstrated in Fig.~\ref{fig:3} for the TSP problems of Fig.~\ref{fig:2}: Transitions that bridge large energy differences occur significantly more frequently in the spiking network, compared to a corresponding non-spiking Boltzmann machine or Gibbs sampling (Fig.~\ref{fig:3}C). In particular, transitions with energy differences beyond $\pm 15$ are virtually absent in Gibbs sampling. This is because groups of neurons that provide strong input to each other (such that all neurons have a high membrane potential $u>15$) are very unlikely to switch off once they have settled into a locally consistent configuration (due to low event rates at high $u$, see Fig.~\ref{fig:3}B). In the spiking network, however, such transitions occur quite frequently, since neurons are ``forced'' to switch off after $\tau$ time units even if they receive strong input $u$, as predicted by Fig.~\ref{fig:3}A. To restore a low-energy state, a neuron (or another neuron in the same WTA circuit with similarly strong input $u$) will likely fire again quickly after reaching the off-state. This gives rise to the observed highly positive and negative energy jumps in the spiking network (see Supplementary material for explanations of further details of Fig.~\ref{fig:3}C). As a consequence of increased state transitions with large energy differences, intermittent transitions into states that leave many problem variables undefined are also more likely to occur in the spiking network (Fig.~\ref{fig:1}C, Fig.~\ref{fig:3}D). 
In order to avoid misunderstandings, we would like to emphasize that such states with undefined problem variables have nothing to do with neurons being in their refractory state, because the state of such neuron is well-defined (with value 1) during that period.

Consistent with the idea that state transitions with large energy differences facilitate exploration, significantly fewer network state changes are needed in the spiking network to arrive at tours with a given desired cost than in the corresponding Boltzmann machine (Fig.~\ref{fig:3}E,F). This demonstrates a surprising advantage of spike-based computation for stochastic search. For the case of Boltzmann machines in discrete time (which is the commonly considered version), the performance difference to spiking neural networks might be even larger. An advantage of spiking neurons for (somewhat artificial) deterministic computations had previously been demonstrated in \cite{Maass:97b}.

Finally, we would like to mention that there exist in other stochastic search methods (e.g. the Metropolis-Hastings algorithm) that are in general substantially more efficient than Gibbs sampling. Hence the preceding results are only of interest if one wants to execute stochastic search by a distributed network in asynchronous continuous time with little or no overhead.

\subsection*{Spike-based internal temperature control}

Most methods that have been proposed for efficient search for low energy states in stochastic systems rely on an additional external mechanism that controls a scaling factor $T$ (``temperature'') for the energy contrast between states, $E_T(\bx) = E(\bx) / T$ (with appropriate renormalization of the stationary distribution). Typically these external controllers lower the temperature $T$ according to some fixed temporal schedule, assuming that after initial exploration the state of the system is sufficiently close to a global (or good local) energy minimum. We propose (principle 4) to exploit instead that a spiking network has in many cases internally information available about the progress of the stochastic search (e.g.~an estimate of the energy of the current state). Making again use of the freedom to use asymmetric synaptic weights, dedicated circuit motifs can internally collect this information and activate additional circuitry that emulates an appropriate network temperature change (Fig.~\ref{fig:1}D). 

More concretely, in order to realize an internal temperature control mechanism for emulating temperature changes of the network energy function according to $E_T(\bx) = E(\bx) / T$ in an autonomous fashion, at least three functional components are required: a) Internally generated feedback signals from circuit motifs reporting on the quality and performance of the current tentative network solution. b) A \emph{temperature control} unit which integrates feedback signals and decides on an appropriate temperature $T$. c) An implementation of the requested temperature change in each circuit motif. 

Both circuits motifs, WTA and OR, can be equipped quite easily with the ability to generate internal feedback signals. The WTA constraint in a WTA circuit is met if exactly one principal neuron is active in the circuit. Hence, the summed activity of WTA neurons indicates whether the constraint is currently met. Similarly, the status of an OR constraint can be read out through an additional status neuron which is spontaneously active but deactivated whenever one of the OR neurons fires. The activity of the additional status neuron then indicates whether the OR constraint is currently violated.

Regarding the temperature control unit, one can think of various smart strategies to integrate feedback signals in order to decide on a new temperature. In the simplest case, a temperature control unit has two temperatures to choose from: one for exploration (high temperature), and for stabilization of good solutions (low temperature). A straightforward way of selecting a temperature is to remain at a moderate to high temperature (exploration) by default, but switch temporarily to low temperature (stabilization) whenever the number of positive feedback signals exceeds some threshold, indicating that almost all (or all) constraints in the circuit are currently fulfilled.

Concretely, such internal temperature control unit can be implemented via a temperature control neuron with a low bias and connection strengths from feedback neurons in each circuit in such a manner that the neuron's firing probability reaches non-negligible values only when all (or almost all) feedback signals are active. When circuits send either positive or negative feedback signals, the connection strengths from negative feedback neurons should be negative and can be chosen in such a manner that non-negligible firing rates are achieved only if all positive feedback but none of the negative feedback signals are active. Whenever such temperature control neuron is active it indicates that the circuit should be switched to the low temperature (stabilization) regime. For details see Supplementary material section 5.

\begin{figure}[!tbp]
  \begin{center}
  \includegraphics[width=\textwidth]{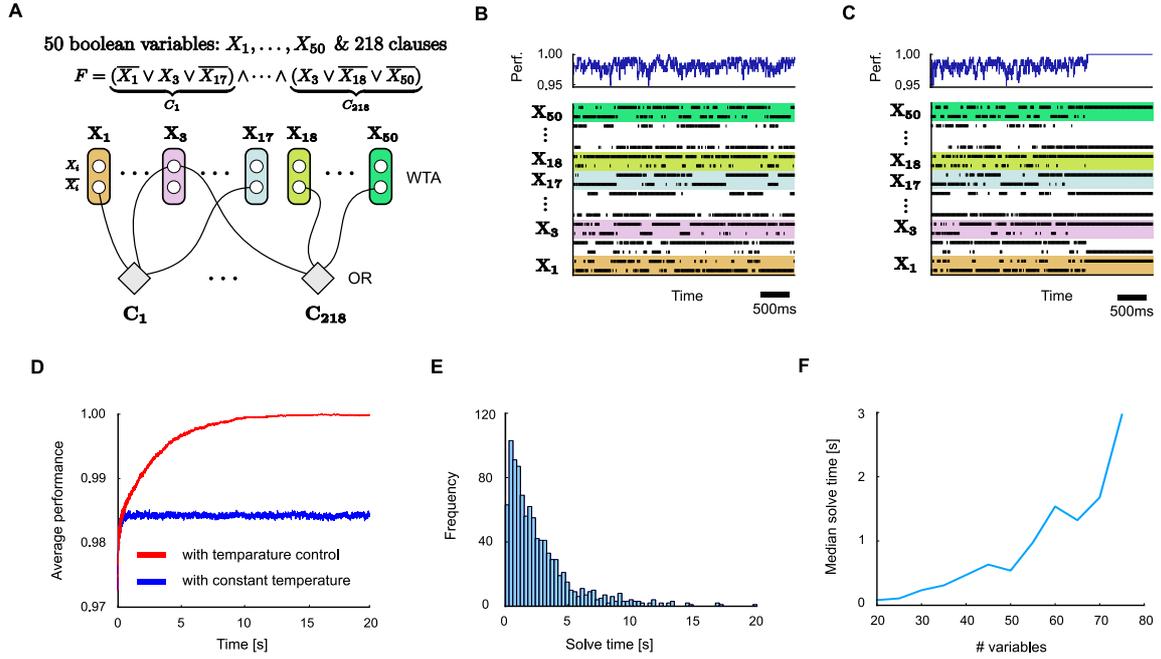}
\caption{Application to 3-SAT. \textbf{A}. Network design. \textbf{B}. Spiking activity in the network (bottom) and quality of problem solution ($\%$ of clauses satisfied) corresponding to each network state. \textbf{C}. Same with internal temperature control circuit motifs added. \textbf{D}. Comparison of quality of solutions represented by networks states $\bx(t)$ in networks with and without internal temperature control (average over 100 runs). \textbf{E}. Distribution of solve times, i.e. the time needed until a satisfying solution is found for the first time, for the Boolean formula from A (with internal temperature control). \textbf{F}. Test of scalability of spike-based design principles to instances of 3-SAT with different numbers of variables (but fixed clause/variable ratio of 4.3). While the network sizes scale linearly, the solve time grows apparently exponentially, as expected for NP-hard problems. Nevertheless, spike-based networks are typically able to solve fairly large hard 3-SAT instances within a few seconds.}
 \label{fig:4}
\end{center}
\end{figure}

\subsection*{Application to the Satisfiability Problem (3-SAT)}

We demonstrate internal temperature control in an application to 3-SAT, another well-studied benchmark task (Fig.~\ref{fig:4}A). 3-SAT is the problem to decide whether a Boolean formula $F$ involving $N$ Boolean variables is satisfiable (or equivalently, whether its negation is not provable), for the special case where $F$ is a conjunction (AND) of clauses (OR's) over 3 literals (i.e., over Boolean variables $X_n$ or their negations $\bar{X}_n$). 
3-SAT is NP-complete, i.e. there are no known methods that can efficiently solve general 3-SAT problems. Large satisfiability problems appear in many practical applications such as automatic theorem proving, planning, scheduling and automated circuit design \cite{biere2009handbook}.

Despite exponential worst-case complexity, many 3-SAT instances arising in practice can be solved quite efficiently by clever (heuristic) algorithms \cite{gomes2008satisfiability}. We are not aware of spiking network implementations for solving satisfiability problems. A class of particularly hard 3-SAT instances can be found in a subset of random 3-SAT problems. In a (uniform) random 3-SAT problem with $N$ Boolean variables and $M$ clauses, the literals of each clause are chosen at random from a uniform distribution (over all possible $2N$ literals). The typical hardness of a random 3-SAT problem is determined to a large extent by the ratio $M/N$ of clauses to variables. For small ratios almost all random 3-SAT problems are satisfiable. For large ratios almost all problems are unsatisfiable. For large problems $N\gg 1$ one observes a sharp phase transition from all-satisfiable to all-unsatisfiable at a crossover point of $M/N\approx 4.26$. For smaller $N$ the transition is smoother and occurs at slightly higher ratios. Problems near the crossover point appear to be particularly hard in practice~\cite{crawford1996experimental}.

\begin{figure}[!tbp]
  \begin{center}
  \includegraphics[width=0.9\textwidth]{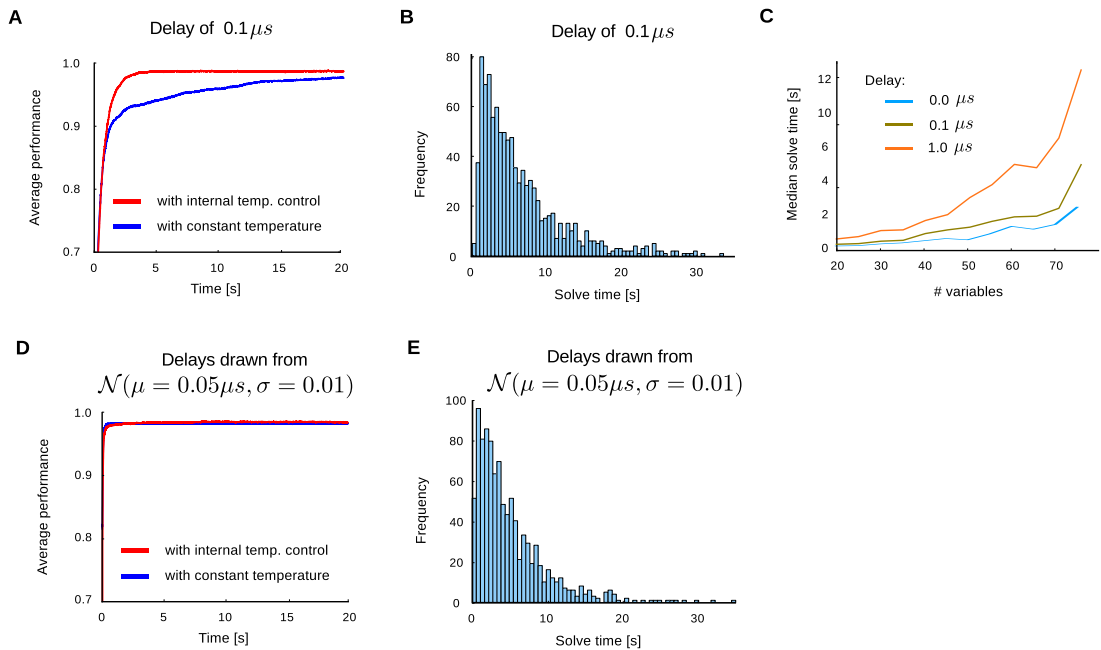}
\caption{Influence of transmission delays between neurons on network performance for 3-SAT. \textbf{A}. Performance with and without internal temperature control (average over 100 runs), with uniform delay $0.1\mu s$. \textbf{B}. Distribution of solve times with uniform delay $0.1\mu s$. \textbf{C}. Comparison of median solve times for different problem sizes, using delays $0/0.1/1\mu s$. \textbf{D}. Network performance with non-uniform delays, where the transmission delay of each connection is drawn from a Gaussian distribution $\mathcal{N}(\mu = 0.05\mu s, \sigma = 0.01)$. \textbf{E}. Distribution of solve times with these randomly drawn delays is similar to that with uniform delays in panel B.} 
\label{fig:5}
 \end{center}
\end{figure}

For hard random instances of 3-SAT, like those considered in Fig.~\ref{fig:4} with a clauses-to-variables ratio 4.3 near the crossover point, typically only a handful of solutions in a search space of $2^{N}$ possible assignments of truth values to the Boolean variables exist. A spike-based stochastic circuit that searches for satisfying value assignments of a 3-SAT instance $F$ can be constructed from WTA- and OR-modules in a straightforward manner (Fig.~\ref{fig:4}A). Each Boolean variable $X_n$ is represented by two neurons $\nu_{n0}$ and $\nu_{n1}$, so that a spike of neuron $\nu_{ni}$ sets the value of $X_n$ to $i$ for a time interval of length $\tau$. A WTA circuit ensures that for most time points $t$ this holds for exactly one of these two neurons (otherwise $X_n$ is undefined at time $t$). An OR-module for each clause of $F$ increases the energy function of the network according to the number of clauses that are currently unsatisfied. An internal spike-based temperature controller can easily be added via additional modules (for details see Supplementary material section 5). It considerably improves the performance of the network (Fig.~\ref{fig:4}C-D), while keeping the total number of neurons linear in the size of the problem instance $F$.

\subsection*{Role of precise timing of spikes}

In spite of the stochasticity of the spiking network, the timing of spikes and subsequent EPSPs plays an important role for their computational function. The simulation results described above were obtained in the absence of transmission delays. Fig.~\ref{fig:5}A-C summarize the impact on performance when uniform delays are introduced into the network architecture of
Fig.~\ref{fig:4}A. For a small delay of 0.1$\mu$s, only a mild performance degradation is observed (compare Fig.~\ref{fig:5}A-C with Fig.~\ref{fig:4}D-F). Computations times remain in the same order of magnitude as with the zero-delay case for delays up to 1$\mu$s (Fig.~\ref{fig:5}C). Higher delays were observed to lead to more significant distortions and substantial performance reduction.

In order to test whether uniformity of delays is essential for the computational performance we also investigated the impact of non-uniform delays, where delays were randomly and independently drawn from a normal distribution with $\mu=0.05\mu s$ and $\sigma=0.01$ truncated at 0 and 0.1$\mu s$ (Fig.~\ref{fig:5}D,E). 
Our results suggest that a variability of delays does not give rise to an additional degradation of performance, only the maximal delay appears to matter.

Altogether we have shown that transmission 
delays of more than 0.1$\mu$s impair the computational performance. Spike transmission within 0.1$\mu$s can easily be achieved in electronic hardware, but not in the brain. On the other hand no brain has apparently been able to solve such hard computational tasks, especially not within seconds -- like our network. One possibility to slow down such computations is to define network states by the set (or sequence) of neurons that fire during a cycle of some background oscillation (as in \cite{JezekETAL:11}). In this way longer transmission delays can be tolerated.

\section*{Discussion}

We have presented a theoretical basis for designing networks of spiking neurons which can solve complex computational tasks in an efficient manner. One new feature of our approach is to design networks of spiking neurons not on the basis of desirable signal cascades and computational operations on the level of neurons and neural codes. Rather, we propose to focus immediately on the network level, where the probability distribution of spike-based states of the whole network (or in alternative terminology: the energy landscape of spike-based network states) defines the conceptual and mathematical level on which the network needs to be programmed (or taught, if one considers a learning approach). We have demonstrated that properties of the energy function of the network can easily be „programmed“ through modular design  on the level of network motifs, that each contribute particular identifiable features to the global energy function. A principled understanding of the interaction between local network motifs and global properties of the energy function of the network is provided by a Modularity Principle (Theorem 1). 
The local network motifs that we consider can be seen as analogues to the classical computing primitives of deterministic digital circuits (Boolean gates) for the design of computationally powerful spike-based circuits with noise.

The resulting spike-based networks differ from previously considered ones in that they use noise as computational resource. Without noise, the network would get stuck in local minima of its energy landscape. A surprising feature of the resulting stochastic networks is that they benefit in spite of the noise from the possibility to use the timing of spikes as a vehicle for encoding information ~~during a computation. But not~ the temporal order of spikes~~ becomes here~ computationally~ relevant, but ~~(almost-)~coincidences of spikes from several neurons within a short time window (in spite of the fact their computations takes place in continuous time, without any clock). This arises from the fact that we define network states (see Fig.~\ref{fig:1}A) by recording which neuron fires within a short time window. 

Finally, we have addressed the question whether there are cases where spike-based computation is faster than a corresponding non-spiking computation. We have shown in Fig.~\ref{fig:3} that this is the case for the TSP. The TSP is a particularly suitable computational task for such comparison, since it can be solved also by a Boltzmann machine with exactly the same architecture. In fact, we have considered for comparison a Boltzmann machine that has in addition the same energy function (or equivalently: the same stationary distribution of network states) as the spiking network. Nevertheless, the sampling dynamics of a spiking network is fundamentally different, since it has an inherent mechanism for bypassing high energy barriers in the search for a state with really low energy. Stochastically spiking neurons change their state more often when their membrane potential is high. As a result, low energy states are visited more often, and hence also sooner, by a spiking network than by a Boltzmann machine or Gibbs sampling. In addition, spiking networks can carry out massively parallel computations without a clock or other organizational overhead. 

\section*{Methods}

\subsection*{Details to the TSP application (Fig.~\ref{fig:2})}

The Traveling Salesman Problem (TSP) is among the most well-known combinatorial optimization problems \cite{cook2011combinatorial} and has been studied intensely for both theoretical and practical reasons: TSP belongs to the class of NP-hard problems, and hence no polynomial-time algorithm is known for solving TSP instances in general. Nevertheless, TSPs arise in many applications, e.g. in logistics, genome sequencing, or the efficient planning of laser positioning in drilling problems \cite{applegate2011traveling}. 
Substantial efforts have been invested in the development of efficient approximation algorithms and heuristics for solving TSP instances in practice. In the Euclidean (planar) case, for example, where movement costs $c_{ij}$ correspond to Euclidean distances between cities in a two-dimensional plane, a polynomial-time approximation scheme (PTAS) exists which is guaranteed to produce approximate solutions within a factor of $(1+\epsilon)$ of the optimal tour in polynomial time \cite{arora1998polynomial}. Various other heuristic algorithms for producing approximate or exact solutions (with typically weaker theoretical support) are often successfully used in practice \cite{applegate2011traveling}. An implementation for solving TSPs with artificial non-spiking neurons was first provided in the seminal paper by Hopfield and Tank \cite{Hopfield:86}. \cite{malaka2000solving} ported their approach to deterministic networks of spiking neurons and reported that such networks found an optimal solution for a planar TSP instance with 8 cities. We are not aware of previous work on solving TSP instances with stochastic spiking neurons.

For a general TSP problem described by a set of $N$ nodes and $N\cdot(N-1)$ edges with associated costs $c_{ij}$ to move from one node to another, the goal is to find a tour with minimum cost which visits each node exactly once. When the cost $c_{ij}$ to move from node $i$ to node $j$ equals the cost $c_{ji}$ from $j$ to $i$ for all pairs of nodes then the problem is called symmetric. If movement costs $c_{ij}$ are Euclidean distances between cities in a two-dimensional plane, the TSP problem is called Euclidean or planar.

Substantial efforts have been invested in the development of efficient approximation algorithms and heuristics for solving TSP instances in practice. In the Euclidean (planar) case, for example, a polynomial-time approximation scheme (PTAS) exists which is guaranteed to produce approximate solutions within a factor of $(1+\epsilon)$ of the optimal tour in polynomial time \cite{arora1998polynomial}. Various other heuristic algorithms with weak theoretical support for producing approximate or exact solutions are often successfully used in practice. One of the most efficient heuristic solvers producing exact solutions for practical TSP problems is \verb|CONCORDE| \cite{applegate2011traveling}. An implementation for solving TSPs with artificial neurons was first provided in the seminal paper of Hopfield and Tank \cite{Hopfield:86}. The neurons in this model were deterministic and analog. Due to deterministic dynamics it was observed that the network often got stuck in infeasible or suboptimal solutions. Various incremental improvements have been suggested to remedy the observed shortcomings \cite{van1989improving, chen1995chaotic}.

\subsubsection*{Network architecture for solving TSP}

In Fig.~\ref{fig:2} we demonstrate an application of our theoretical framework for stochastic spike-based computation to the TSP. 
To encode the node (city) visited at a given step of a tour in a TSP problem consisting of $N$ nodes, a discrete problem variable with $N$ possible values is needed. To encode a tour with $N$ steps, $N$ such problem variables are required. More precisely, for efficiency reasons we consider in our approach a relaxed definition of a tour: a tour for an $N$-node problem consists of $N' = N + N_{\textnormal{resting}}$ steps (and problem variables). A tour is valid if each node is visited at least once. Since there are more steps than nodes, some nodes have to be visited twice. We require that this can only occur in consecutive steps, i.e. a the salesman may remain for at most one step in a city before he must move on. We observed that using such relaxed definition of a tour with $N_{\textnormal{resting}}$ additional ''resting`` steps may considerably improve the efficiency of the stochastic search process. 

To solve a TSP problem one then needs to consider three types of constraints: 
\begin{enumerate}
\item[(a)] Each of the $N'$ problem variables should have at any moment a uniquely defined value assigned to it.
\item[(b)] Each value from the problem variable domain $\{1,\dots, N\}$ must appear at least once in the set of all problem variables (each node has to be visited at least once). At the same time only neighboring problem variables, i.e. those coding for consecutive steps, may have the identical values (allowing for ''resting`` steps).
\item[(c)] The penalty that two consecutive problem variables appear in a given configuration, i.e. a particular transition from node $i$ to some other node $j$, must reflect the traveling cost between the pair of nodes. 
\end{enumerate}

In the spiking network implementation, each step (problem variable) $n \in \{1, \dots N'\}$ is represented by a set of $N$ principal neurons, $\nu_{n1}, \dots, \nu_{nN}$, i.e.~each principal neuron represents one of the $N$ nodes (cities). A proper representation of problem variables (and implementation of constraints (a)) is ensured by forming a WTA circuit from these $N$ principal neurons which ensures that most of the time only one of these principal neurons is active, as described in Supplementary material section 3.2: The WTA circuit shapes the energy function of the network such that it decreases the energy (increases probability) of states where exactly one of the principal neurons in the WTA is active, relative to all other states. If the WTA circuits are setup appropriately, most of the time exactly one principal neuron in each WTA will be active. The identity of this active neuron then naturally codes for the value of the associated problem variable. Hence, one can simply read out the current assignment of values to the problem variables from the network principal state based on the activity of principal neurons within the last $\tau$ time units.

The parameters of the WTA circuits must be chosen appropriately to ensure that problem variables have a properly defined value most of the time. To achieve this, the inhibitory auxiliary neuron is equipped with a very low bias $b_{inh}$, such that the inhibition and therefore the WTA mechanism is triggered only when one of the principal neurons spikes. The inhibitory neuron is connected to all principal neurons associated with the considered problem variable through bidirectional connections $w_{WTA}$ and $w_{exc}$ (from and to the inhibitory neuron, respectively). The inhibition strength $w_{WTA}$ should be chosen strong enough to prevent all principal neurons in the WTA circuit from spiking (to overcome their biases) and $w_{exc}$ should be set strong enough so that it activates inhibition almost immediately before other principal neurons can spike. At the same time the biases of the principal neurons $b_{WTA}$ should be high enough to ensure sufficient activation of principal neurons within the WTA circuit so that the corresponding problem variable is defined most of the time. To enforce that a specific problem variable assumes a particular value one can modulate the biases of the associated principal neurons and thereby decrease the energy of the desired problem variable value. We used this to set the value of the first problem variable $n=1$ to point to the first node (city). In particular, the bias of the principal neuron $\nu_{11}$ is set to $b_P$ (should be set high enough to ensure exclusive activity of that neuron) and biases of all other principal neurons in the first WTA circuit are set to $b_N$ (should be set low enough to prevent neurons from being active).

The implementation of constraints (b) requires that all problem variables have different values except if they are neighboring variables. This is realized by connecting bidirectionally all principal neurons which code for the same value/node in different WTA circuits with strong negative connection $w_{unique}$, except if WTA circuits (or corresponding problem variables) are neighboring. This simply increases the energy (decreases probability) of the network states where the principal neurons coding for the same value in different WTAs are co-active except if they are associated with neighboring problem variables (WTAs). 

Finally, constraints (c) are implemented by adding synaptic connections between all pairs of principal neurons located in neighboring problem variables (except between principal neurons coding for the same value). A positive connection between two principal neurons increases the probability that these two neurons become co-active (the energy of corresponding network states is decreased), and vice versa. Hence, when movement costs between two nodes (cities) $i$ and $j$ at steps $n$ and $n+1$, respectively, are low, the corresponding neurons should be linked with a positive synaptic connection in order to increase the probability of them becoming active together. We chose to encode weights such that they reflect the relative distances between cities (nodes). To calculate the weights we normalize all costs of the TSP problem with respect to the maximum cost, yielding normalized costs $\tilde{c}_{ij} \in [0,1]$. Then we rescale and set the synaptic connections between neuron $\nu_{ni}$ and $\nu_{(n+1)j}$ according to 
\begin{align}
w_{ni,(n+1)j}=w_{(n+1)j,ni} = w_{offset}+(1-\tilde{c}_{ij})*w_{scale} 
\end{align}
for $n\in \{1, \dots N'-1\}$. The connections between step $N'$ and step $1$ are set in an analogous fashion. As a result of this encoding, the energy landscape of the resulting network assigns particularly low energies to network states representing low-cost tours. 

The proposed neural network architecture for solving TSP results in a ring structure (as the last and the first problem variable are also connected). The architecture requires $(N+1)(N+N_{\textnormal{resting}})$ neurons (including inhibitory neurons in WTA circuits) and $N(N+N_{\textnormal{resting}})( 2N + N_{\textnormal{resting}}-2)$ number of synapses.

\subsubsection*{Parameters and further details to the TSP application}

The planar TSP in Fig.~\ref{fig:2}A is taken from \verb|http://www.math.uwaterloo.ca/tsp/world/countries.html|. The 38 nodes in the problem correspond to cities in Djibouti, and movement costs are Euclidean distances between cities. To solve the problem we used the following setup: PSP length $\tau=10e-3$ and refractory period of $10$ms for all neurons, $b_{WTA}=-0.45$, $b_{P}=100$, $b_{N}=-100$, $b_{inh}=-10$, $w_{WTA}=-100$, $w_{exc}=100$, $w_{unique}=-14.7$, $w_{scale}=19.4$, $w_{offset}=-5$ and $N_{\textnormal{resting}}=7$, resulting in the neural network consisting of 1755 neurons. The value of the first variable (the first step) was fixed to the first city.
For the asymmetric TSP problem, in order to facilitate comparison we chose a similarly sized problem consisting of 39 cities from TSPLIB, \verb|http://comopt.ifi.uni-heidelberg.de/software/TSPLIB95/|, file \verb|ftv38.atsp.gz|.
To solve this asymmetric TSP problem we used the same architecture as above, with slightly different parameters:
$b_{WTA}=1.3$, $w_{unique}=-14.1$, $w_{offset}=-7.9$, $w_{scale}=20.8$, $N_{\textnormal{resting}}=8$, resulting in a neural network consisting of 1880 neurons.

Reading out the current value assignment of each problem variable is done based on the activity of the principal neurons which take part in the associated WTA circuit. The performance of the network is calculated at any moment as the ratio of the optimal tour length and the current tour length represented by the network. In order for the currently represented tour to be valid all variables have to be properly defined and each value (city) has to appear at least once. Although this is not \emph{always} the case, exceptions occur rarely and last for very short periods (in the order of ms) and therefore are not visible in the performance plots.

\subsection*{Details to the 3-SAT application (Fig.~\ref{fig:4})}

3-SAT is the problem of determining if a given Boolean formula in conjunctive normal form (i.e.~conjunctions of disjunctive clauses) with clauses of length 3 is satisfiable. 3-SAT is NP-complete, i.e. there are no known methods that can efficiently solve general 3-SAT problems. The NP-completeness of 3-SAT (and general SATISFIABILITY) was used by Karp to prove NP-completeness of many other combinatorial and graph theoretical problems (Karp's 21 problems,  \cite{karp1972reducibility}). Large satisfiability problems appear in many practical applications such as automatic theorem proving, planning, scheduling and automated circuit design \cite{biere2009handbook}.

Despite exponential worst-case complexity, many problems arising in practice can be solved quite efficiently by clever (heuristic) algorithms. A variety of algorithms have been proposed \cite{gomes2008satisfiability}. Modern SAT solvers are generally classified in complete and incomplete methods. Complete solvers are able to both find a satisfiable assignment if one exists or prove unsatisfiability otherwise. Incomplete solvers rely on stochastic local search and thus only terminate with an answer when they have identified a satisfiable assignment but cannot prove unsatisfiability in finite time. The website of the annual SAT-competition, \verb|http://www.satcompetition.org/|, provides an up-to-date platform for quantitative comparison of current algorithms on different subclasses of satisfiability problems.

Particularly hard problems can be found in a subset of random 3-SAT problems. In a (uniform) random 3-SAT problem with $N$ Boolean variables and $M$ clauses, the literals of each clause are chosen at random from a uniform distribution (over all possible $2N$ literals). The typical hardness of a random 3-SAT problem appears to be determined to a major extent by the ratio $M/N$ of clauses to variables. For small ratios almost all random 3-SAT problems are satisfiable. For large ratios almost all problems are unsatisfiable. For large problems $N\gg 1$ one observes a sharp phase transition from all-satisfiable to all-unsatisfiable at a crossover point of $M/N\approx 4.26$. For smaller $N$ the transition is smoother and occurs at slightly higher ratios. Problems near the crossover point appear to be particularly hard in practice \cite{crawford1996experimental}.

\subsubsection*{Network architecture for solving 3-SAT}

For the demonstration in Fig.~\ref{fig:4} we consider satisfiable instances of random 3-SAT problems with a clause to variable ratio of 4.3 near the crossover point. Based on our theoretical framework a network of stochastic spiking neurons can be constructed which automatically generates valid assignments to 3-SAT problems by stochastic search (thereby implementing an incomplete SAT solver). The construction based on WTA and OR circuits is quite straightforward: each Boolean problem variable of the problem is represented by a WTA circuit with two principal neurons (one for each truth value). Analogous to the TSP application, the WTA circuits ensure that most of the time only one of the two principal neurons in each WTA is active. As a result, most of the time a valid assignment of the associated problem variables can be read out.

Once problem variables are properly represented, clauses can be implemented by additional OR circuits. First, recall that, in order to solve a 3-SAT problem, all clauses need to be satisfied. A clause is satisfied if at least one of its three literals is true. In terms of network activity this means that at least one of the three principal neurons, which code for the involved literals in a clause, needs to be active. In order to encourage network states in which this condition is met, one may use an OR circuit motif which specifically increases the energy of network states where none of the three involved principal neurons are active (corresponding to assignments where all three literals are false). Such application of an OR circuit will result in an increased probability for the clause to be satisfied. If one adds such an OR circuit for each clause of the problem one obtains an energy landscape in which the energy of a network state (with valid assignments to all problem variables) is proportional to the number of violated clauses. Hence, stacking OR circuits implicitly implements the conjunction of clauses.

The described network stochastically explores possible solutions to the problem. However, in contrast to a conventional computer algorithm with a stopping criterion, the network will generally continue the stochastic search even after  it has found a valid solution to the problem (i.e.~it will jump out of the solution again). To prevent this, based on feedback signals from WTA and OR circuits it is straightforward to implement an internal temperature control which puts the network in a low temperature regime once a solution has been found (based on feedback signals). This mechanism basically locks the network into the current state (or a small neighboring region of the current state), making it very unlikely to escape. The mechanism is based on a internal temperature control neuron which receives feedback signals from WTA and OR circuits (signaling that all the constraints of the problem were met) and which activates additional circuits (copies of OR circuits) that effectively put the network in the low temperature regime.

\subsubsection*{Parameters and further details for the 3-SAT application}

Each OR circuit is constructed by adding two auxiliary neurons, I and II, with biases of $0.5B$ and $-3.5B$, respectively, where $B$ is some constant. Both auxiliary neurons connect to those principal neurons which code for a proper value of the literals in the clause (in total to 3 neurons), with bidirectional connections $w_{OR}$ and $-B$ (from and to the auxiliary neuron I, respectively), and $-w_{OR}$ and $B$ (from and to the auxiliary neuron II, respectively). Additionally, the auxiliary neuron I connects to the auxiliary neuron II with the strength $3B$. Here the strength of incoming connection from principal neurons and the bias of auxiliary neuron I are set such that its activity is suppressed whenever there is at least one of three principal neurons active, while for auxiliary neuron II are set such that it can be activated only when auxiliary neuron I and at least one of three principal neurons are active together (see OR circuit motif for more details).

The internal temperature control mechanism is implemented as follows. The regime of low temperature is constructed by duplicating all OR circuits - so by adding for each clause two additional auxiliary neurons III and IV which are connected in the same way as the auxiliary neurons I and II(they target the same neurons and are also connected between themselves in the same manner) but with different weights: $w_{OR2}$ and $-B$ (from and to auxiliary neuron), and $-w_{OR2}$ and $B$(from and to auxiliary neuron), for the III and IV auxiliary neuron, respectively. The use of the same OR motif ensures the same functionality, which is activated when needed in order to enter the regime of low temperature and deactivated to enter again regime of high temperature. The difference in connections strength between $w_{OR1}$ and $w_{OR2}$ determines the temperature difference between high and low temperature regimes. In addition the biases of additional auxiliary neurons are set to $-0.5B$ and $-6.5B$ (III and IV aux. neuron), so that they are activated (i.e.~functional) only when a certain state (the solution) was detected. This is signaled by the global temperature control neuron with bias $b_{glob}$ that is connected to both additional auxiliary neurons of all clauses with connection strengths $B$ and $3B$ to the III and IV auxiliary neuron, respectively, so that once the global temperature control neuron is active the additional auxiliary neurons behave exactly as auxiliary neurons I and II. Additionally, the global temperature control neuron is connected to every other principal neuron with connection strength $w_{glob}$, which mimics the change in temperature regime in WTA circuits. The global neuron is active by default, which is ensured by setting the bias $b_{glob}$ sufficiently high, but is deactivated whenever one of the status neurons, which check if a certain clause is not satisfied, is active (\emph{not OK} signals). There is one status neuron for each clause, with bias set to $-2.5B$, where each one of them receives excitatory connections of strength $B$ from all neurons not associated with that particular clause. Therefore, if all problem variables that participate in the clause are set to the wrong values, this triggers the status neuron which reports that the clause is not satisfied. This automatically silences the global neuron signaling that the current network state is not a valid solution. Note that implementation of such internal temperature control mechanism results with inherently non symmetric weights in the network.

For described architecture of a spiking neural network the total number of neurons needed is $3N+2M$ ($2+1$ per WTA circuit, and $2$ per OR circuit), while the number of connections is $4N + 13M$. Notably, both the number of neurons and the number of synapses depend linearly on the number of variables $N$ (the number of clauses linearly depends on the number of variables if problems with some fixed clauses-to-variables ratio are considered). To implement in addition described internal temperature control mechanism one needs additional $3M+1$ neurons and $2N+20M$ synapses.

The architecture described above was used in simulations with the following parameters: $\tau=10e-3$ and refractory period of $10$ms for all neurons except for the global neuron which has $\tau=9e-3$ and refractory period of $9$ms, $b_{WTA}=2$, $b_{inh}=-10$, $b_{glob}=10$, $B=40$, $w_{WTA}=-100$, $w_{exc}=100$, $w_{OR1}=2.5$, $w_{OR2}=10$, with rectangular PSPs of $10$ms duration without transmission delays for all synapses except for the one from the global neuron to the additional auxiliary neurons where the PSP duration is $11$ms. The network which solves the considered Boolean problem in Fig.~\ref{fig:4}A consisting of 50 variables and 218 clauses has 586 neurons.

To calculate the performance of a solution at any point in time we use as a performance measure the ratio between the number of satisfied clauses and total number of clauses. If none of the variables which take part in a clause are properly defined then that clause is considered unsatisfied. As a result, this performance measure is well-defined at any point in time.

For the analysis in Fig.~\ref{fig:4}F of problem size dependence we created random 3-SAT problems of different sizes with clause-to-variable ratio of 4.3. To ensure that a solution exists, each of the created problems was checked for satisfiability with zhaff, a freely available (complete) 3-SAT solver. 

Finally, note that the proposed architecture can be used in analogues way in order to solve k-SAT problems, where each clause is composed of $k$ literals. In this case the only change concerns OR circuit motifs which now involve for each clause $k$ associated principal neurons.

\subsection*{Software used for simulations}

All simulations were performed in NEVESIM, an event-based neural simulator \cite{PecevskiETAL:14}.
The analysis of simulation results was performed in Python and Matlab.

\bibliography{CSP.bib}

\section*{Acknowledgments}

We would like to thank Dejan Pecevski for givinig us advance access to the event-based neural simulator NEVESIM \cite{PecevskiETAL:14}, that we used for all our simulations. We thank Robert Legenstein, Gordon Pipa and Jason MacLean for comments on the manuscript. The work was partially supported by the European Union project \#604102 (Human Brain Project).

\section*{Author contributions}
Conceived and designed the experiments: Z.J., S.H. and W.M. Performed the experiments: Z.J. Wrote the paper: Z.J., S.H. and W.M. Theoretical analysis: S.H. and W.M.

\section*{Competing Financial Interests statement}
The authors declare no competing financial interests.

\end{document}


\title{Supplementary material:\\ A theoretical basis for efficient computations 
\\with noisy spiking neurons}
\author{\small Zeno Jonke$^\dagger$, Stefan Habenschuss$^\dagger$, Wolfgang Maass\footnote{Corresponding author: maass@igi.tugraz.at. $^\dagger$ These authors contributed equally to this work. }\\\small 
Institute for Theoretical Computer Science, Graz University of Technology, Austria}

\maketitle

\section{Stochastic neuron model}

Neurons are modeled as simple stochastic point neurons with absolute refractory period $\tau$. When not in a refractory state, neuron $k$ spikes at an instantaneous firing rate which depends exponentially on the membrane potential $u_k(t)$ given by equation \prettyref{eq:uk}\footnote{We use prefix M to refer to the main text material.}, according to,
\begin{align}
\lim_{\delta t \rightarrow 0} p(\,\textnormal{neuron } k \textnormal{ fires in } (t, t+\delta t]\,)/\delta t = \rho_k(t) = \frac{1}{\tau} \exp(u_k(t))\;\;,
\end{align}
with $\tau=10$ms unless otherwise stated. An exponential dependence of a neuron's firing probability on the membrane potential has been suggested by \cite{JolivetETAL:06} based on a fit to experimental data. Similar stochastic neuron models have been suggested by \cite{truccolo2005point, BuesingETAL:11}. 

Note that an exponential firing function is required for correct sampling when a constant ``off-rate'' is assumed (corresponding to a constant PSP length which does not depend on the pre-synaptic membrane potential). In the discrete-time variant of neural sampling \cite{BuesingETAL:11}, the resolution (how many discrete time steps constitute a refractory period) determines whether the canonical activation function resembles more a sigmoid or an exponential function. In discrete time implementations of neural sampling it may be thus expected that a variety of intermediate behaviors between Gibbs sampling and continuous-time neural sampling is found at different resolutions.

\section{Details to Principle 1: Stationary distributions and energy functions}

\subsection{Network states}

The state $x_{k}(t)$ of a principal neuron $k$ at time $t$ is defined as,
\begin{align}
x_k(t) = \begin{cases}
          1, & \textnormal{if neuron k fired within } (t-\tau, t]\;\;,\\
          0, & \textnormal{otherwise}\;\;,
         \end{cases}\label{seq:princxk}
\end{align}

\noindent where $\tau$ is a brief time window corresponding to the duration of a PSP. The state vector of all principal neurons (the \emph{principal network state}) is denoted by $\mathbf{x}(t)=\left(x_1(t), \dots, x_N(t)\right)$. Unless otherwise stated, the term \emph{network state} refers to the principal network state.

The state $\xi_{m}(t)$ of an auxiliary neuron $m$ is defined in an analogous manner to equation \eqref{seq:princxk}. The state vector of all auxiliary neurons is written as $\boldsymbol{\xi}(t)=\left(\xi_1(t), \dots, \xi_M(t) \right)$. The \emph{full network state} is given by,
\begin{align}
\left(\mathbf{x}(t), \boldsymbol{\xi}(t)\right)=\left(x_1(t), \dots, x_N(t), \xi_1(t), \dots, \xi_M(t) \right) 
\end{align}

Similar notions of network state have been suggested by a number of experimental and theoretical papers \cite{schneidman2006weak,BerkesETAL:11,BuesingETAL:11,Pecevski2011} and \cite{HabenschussETAL:13}.

\subsection{Convergence to stationary distribution}

Under mild conditions, activity in a general spiking network with noise can be theoretically guaranteed to converge exponentially fast to a unique stationary distribution $p(\mathbf{x}, \boldsymbol{\xi})$ of full network states \cite{HabenschussETAL:13}, regardless of initial network conditions. In the context of the stochastic neuron model , equations \prettyref{eq:neuronmodelrho}-\prettyref{eq:uk}, it can be easily verified that the theoretical conditions for convergence are fulfilled if all weights are finite. Clearly, throughout the paper this condition is met. Exponentially fast convergence to a unique \emph{marginal} distribution $p(\mathbf{x})$ over principal network states is a simple corollary that follows from the convergence to a unique joint distribution $p(\mathbf{x}, \boldsymbol{\xi})$.

\subsection{Energy functions}

In analogy with statistical physics \cite{plischke2006equilibrium}, we define the energy function $E(\mathbf{x})$ of a network of spiking neurons with a unique stationary distribution $p(\mathbf{x})$ of principal network states $\mathbf{x}$ as 
\begin{align}
 E(\mathbf{x}) = - \log p(\mathbf{x}) + C\;\;,
\end{align}
with an arbitrary constant $C$. The stationary distribution $p(\mathbf{x})$ can then be expressed as,
\begin{align}
p(\mathbf{x}) = \frac{e^{-E(\mathbf{x})}}{\sum_{\mathbf{x}'} e^{-E(\mathbf{x}')}}\;\;.
\end{align}

Note that according to this definition, energies are defined only up to a constant (a global shift applied to all states). To indicate that two energy functions are identical except for a constant shift we use the notation $E_1(\mathbf{x}) \triangleq E_2(\mathbf{x})$, i.e.~
\begin{align}
E_1(\mathbf{x}) \triangleq E_2(\mathbf{x})  \quad \Leftrightarrow\quad \exists{C}{\in}{\mathbb{R}} \; \forall{ \mathbf{x}} \; \left( E_1(\mathbf{x}) = E_2(\mathbf{x}) + C \right) \;\;.
\end{align}

\section{Details to Principle 2: Shaping the energy function through circuit motifs}

A key theoretical question is how the energy function $E(\mathbf{x})$ (or equivalently $p(\mathbf{x})$) over principal network states $\mathbf{x}$ depends on the parameters of a network, in particular on synaptic weights $w_{kl}$ and neuronal excitabilities $b_k$ among principal neurons, as well as on auxiliary circuits connected to the principal neurons.
Previous work had shown that pair-wise symmetric connections between neurons map onto second-order dependencies between variables \cite{BuesingETAL:11}. \cite{Pecevski2011} demonstrated in addition how more complex dependencies can be encoded through the use of pre-processing circuits in the context of probabilistic inference. 

Here we consider how in addition to second-order dependencies, common higher-order constraints of hard computational problems can be encoded through the use of simple auxiliary circuit motifs, in a manner suitable for modularity and large-scale circuit design.

\subsection{Conditions for modularity}

To facilitate systematic design of complex energy landscapes, we would like to find a basic set of auxiliary circuit motifs which can be combined in arbitrarily rich ways with predictable outcomes. A particularly desirable feature to aim for is linear modularity, such that the energy contribution to the energy landscape of each circuit motif is independent of the presence of other circuit motifs. 

The starting point for the following results is the Neural Computability Condition (NCC) from \cite{BuesingETAL:11}, which provides a sufficient condition for a network of model neurons given by equation \prettyref{eq:neuronmodelrho} to sample from some desired distribution $p(\mathbf{x})$. The NCC requires that the membrane potential of each neuron $k$ obeys,
\begin{align}
 u_k(t) = \log \frac{p(x_k = 1|\mathbf{x}_{\setminus k} (t))}{p(x_k = 0|\mathbf{x}_{\setminus k}(t))}\;\;.\label{seq:NCC2}
\end{align}
where $\mathbf{x}_{\setminus k}(t)$ denoting the current state vector of all principal neurons
excluding neuron $k$. In terms of a desired energy function $E(\mathbf{x})$, the NCC can be reformulated as
\begin{align}
 u_k(t) = E(x_k = 0, \mathbf{x}_{\setminus k}(t)) - E(x_k = 1, \mathbf{x}_{\setminus k}(t))\;\;,\label{seq:NCCE}
\end{align}
where we use the simplified notation $E\left(x_k = \cdot, \mathbf{x}_{\setminus k}(t)) := E(x_1(t), \dots, x_{k-1}(t), \cdot, x_{k+1}(t), \dots, x_N(t)\right)$.

In the absence of auxiliary neurons/circuits it was shown by \cite{BuesingETAL:11} that, if all synaptic connections among principal neurons are symmetric ($w_{kl}=w_{lk}$), a network $\mathcal{N}$ of principal neurons will sample from a Boltzmann distribution with energy function given by equation \prettyref{eq:boltze}. Suppose that a set of auxiliary circuits $\mathcal{I}$ is added (and connected) to such a principal network. Then, due to linearity of membrane integration ,see equation \prettyref{eq:uk}, the membrane potential of a principal neuron $k$ in the presence of such auxiliary circuits can be written as equation ~\prettyref{eq:ukaux}.

The energy contribution that each auxiliary circuit $C_i$ makes to the energy function $E(\mathbf{x})$ of principal neurons, however, may be arbitrarily complex. In particular, it may depend in non-trivial ways on the presence and detailed structure of other auxiliary circuits and synaptic weights and biases of the principal network. Under appropriate conditions, however, the energy contribution of each auxiliary circuit becomes linear and independent of the remaining network (Theorem~1).

Theorem~1 suggests that auxiliary circuits should be constructed in a highly specific manner to support modularity. In particular, condition given by equation \prettyref{eq:deltaudeltae} states that auxiliary circuit contributions to the membrane potential of a principal neuron $k$ should be basically memoryless and reflect a specific function of the current state of the remaining network, $\mathbf{x}_{\setminus k}(t)$. Note that this function (the right-hand side of equation \prettyref{eq:deltaudeltae}) has a very intuitive interpretation: a circuit $C_i$ should inform each principal neuron $k$ about the currently expected drop in the energy function $U_i$ that can be achieved by a spike of neuron $k$ (i.e.~a switch from $x_k=0$ to $x_k=1$).
\\\\
\textbf{Proof of Theorem 1: } If equation \prettyref{eq:deltaudeltae} holds for all $C_i$ then the membrane potential of a principal neuron $k$ in the presence of some subset of auxiliary circuits  $C_i,i \in  \mathcal{I}$, is given at time $t$ by,
\begin{align}
u_{k, \mathcal{I}}(t) &= b_k + \sum_{l=1}^N w_{kl}\,x_l(t) + \sum_{i \in \mathcal{I}} \left[U_i\left(x_k=0, \mathbf{x}_{\setminus k}(t)\right) - U_i\left(x_k=1, \mathbf{x}_{\setminus k}(t)\right)\right]\;\;.\label{seq:umemtotal}
\end{align}
To verify that the network has stationary distribution
\begin{align}
p(\bx) \propto e^{-E_{\mathcal{N}}(\mathbf{x}) - \sum_{i \in  \mathcal{I}} U_i(\mathbf{x})}\label{seq:statio}
\end{align}
we plug into the NCC given by equation \eqref{seq:NCC2} from \cite{BuesingETAL:11},
\begin{align}
\log\frac{p(x_k=1|\mathbf{x}_{\setminus k})}{p(x_k=0|\mathbf{x}_{\setminus k})} &= \log\frac{p(x_k=1, \mathbf{x}_{\setminus k})}{p(x_k=0, \mathbf{x}_{\setminus k})} \\
&= \log p(x_k=1, \mathbf{x}_{\setminus k}) - \log p(x_k=0, \mathbf{x}_{\setminus k})\\
 &= -E_{\mathcal{N}, \mathcal{I}}(x_k=1, \mathbf{x}_{\setminus k})  + E_{\mathcal{N}, \mathcal{I}}(x_k=0, \mathbf{x}_{\setminus k})\\
& = -E_{\mathcal{N}}(x_k=1, \mathbf{x}_{\setminus k})  + E_{\mathcal{N}}(x_k=0, \mathbf{x}_{\setminus k}) \\
&\quad\quad + \sum_{i\in\mathcal{I}} [-U_{i}(x_k=1, \mathbf{x}_{\setminus k})  + U_{i}(x_k=0, \mathbf{x}_{\setminus k})]\\
&= b_k + \sum_{l=1}^N w_{kl} x_l + \sum_{i\in\mathcal{I}} [-U_{i}(x_k=1, \mathbf{x}_{\setminus k})  + U_{i}(x_k=0, \mathbf{x}_{\setminus k})]\;\;.
\end{align}
Thus, a network with membrane dynamics given by equation \eqref{seq:umemtotal} meets the NCC for the distribution in equation \eqref{seq:statio}. As a result, equation \eqref{seq:statio} must be the unique stationary distribution of the network, and the energy function is given by $E_{\mathcal{N}}(\mathbf{x}) + \sum_{i \in  \mathcal{I}} U_i(\mathbf{x})$. 
\\$\Box$\\

Note that, in contrast to neural sampling theory \cite{BuesingETAL:11}, Theorem~1 is concerned with the distribution over a subset of all neurons (the principal neurons $\mathbf{x}$), i.e. the marginal distribution $p(\mathbf{x})$ after integrating out all auxiliary variables $\boldsymbol{\xi}$. A concrete application of Theorem~1 is the design of auxiliary circuit motifs which \emph{approximate} equation \prettyref{eq:deltaudeltae} in practice, as described below for the WTA and OR motifs.

\subsection{WTA circuit motif}\label{subsec:motif}

The WTA circuit motif consists of a single auxiliary neuron which is reciprocally connected to some subset \mbox{$\mathcal{K} \subseteq \{1, \dots N\}$} of principal neurons (Fig.~\ref{fig:1}B, top left). The goal of the WTA motif is to achieve that most of the time \emph{exactly} one neuron in $\mathcal{K}$ is active: whenever one principal neuron spikes it activates the (inhibitory) auxiliary neuron which suppresses all other principal neurons. 

More precisely, in terms of energies, the goal of the WTA motif is to increase the energies of all network states with more or less than one active neuron in $\mathcal{K}$. This can be achieved in two steps. First, the energy of all network states where more than one neuron in $\mathcal{K}$ is active is increased.
We found that this can be robustly achieved by a single inhibitory neuron which receives strong excitatory connections from $\mathcal{K}$, and sends strong inhibitory connections back to $\mathcal{K}$ (with some weight $w_{\textnormal{WTA}}\ll 0$). The inhibitory neuron should have a low bias such that it only fires when one of the principal neurons is active. Second, the energy of states where no neuron in $\mathcal{K}$ is active is raised. This can be done most easily by raising the biases of all neurons in $\mathcal{K}$ by some constant $b_{\textnormal{WTA}}$  with $0 < b_{\textnormal{WTA}} < -w_{\textnormal{WTA}}$. Alternatively, this could in principle also be achieved by an additional auxiliary neuron which is constantly active and makes excitatory connections to all neurons in $\mathcal{K}$. 

The design of the described implementation of the WTA circuit motif was based on the Modularity Principle (Theorem~1). This can be seen if one considers the desired energy function
\begin{align}
U_{\textnormal{WTA}[\mathcal{K}]} (\mathbf{x}) = \begin{cases}
	b_{\textnormal{WTA}}\;, & \textnormal{if }\sum_{k \in \mathcal{K}} x_k = 0\;,\\
	0\;, & \textnormal{if }\sum_{k \in \mathcal{K}} x_k = 1\;,\\
	(-w_{\textnormal{WTA}} - b_{\textnormal{WTA}}) \cdot (- 1 + \sum_{k \in \mathcal{K}} x_k )\;, & \textnormal{if }\sum_{k \in \mathcal{K}} x_k > 1\;.
                                                 \end{cases}\label{seq:WTAU}
\end{align}
Theorem~1 provides a concrete guideline for the design of an auxiliary circuit implementing this energy function. In particular, according to equation \prettyref{eq:deltaudeltae} the ideal membrane potential contribution of the auxiliary circuit to principal neuron $k$, $\Delta u_{k,\textnormal{WTA}[\mathcal{K}]}(t)$, for implementing equation \eqref{seq:WTAU} in a modular manner is given by,
\begin{align}
 \Delta u_{k,\textnormal{WTA}[\mathcal{K}]}(t) &= U_{\textnormal{WTA}[\mathcal{K}]} (x_k=0, \mathbf{x}_{\setminus k}) - U_{\textnormal{WTA}[\mathcal{K}]} (x_k=1, \mathbf{x}_{\setminus k})\\
&= 
\begin{cases}
b_{\textnormal{WTA}}\;, &\sum_{l \in \mathcal{K}\setminus k} x_l(t) = 0\;,\\
b_{\textnormal{WTA}}-w_{\textnormal{WTA}} \;, &\sum_{l \in \mathcal{K}\setminus k} x_l(t) > 0\;.
\end{cases}
\end{align}

This membrane potential contribution is closely approximated by the described WTA circuit implementation: Regardless of the network state, there is a bias term $b_{\textnormal{WTA}}$ for each principal neuron $k\in\mathcal{K}$. As soon as one (or more) of these neurons fire, this triggers the auxiliary inhibitory neuron, which then strongly inhibits all competitors in $\mathcal{K}$ with weight $w_{\textnormal{WTA}}$. The nature of the approximation lies mainly in the delay between the onset of activity of a winner and the onset of inhibition at the remaining principal neurons (due to stochastic firing of the auxiliary neuron).

\subsection{OR circuit motif}\label{subsec:OR-motif}

The OR circuit motif consists of two auxiliary neurons $I$ and $II$ reciprocally connected to some subset \mbox{$\mathcal{K} \subseteq \{1, \dots N\}$} of principal neurons (Fig.~\ref{fig:1}B, top right). The purpose of the OR motif is to ensure that most of the time \emph{at least one} neuron in $\mathcal{K}$ is active. In essence, the auxiliary OR circuit motif remains silent as long as this OR-condition is met and at least one neuron in $\mathcal{K}$ is active. The motif is activated whenever it is detected that no neuron in $\mathcal{K}$ is active. The auxiliary circuit then excites the principal neurons until one of them fires.

At first sight, it may appear that this basic functionality should require only one auxiliary neuron, neuron~$I$, which is inhibited by all principal neurons but starts firing immediately upon disinhibition when no principal neuron is active. Neuron $I$ then keeps firing and exciting the principal neurons (with synaptic weight $w_{OR}$) until the OR-condition is restored.

However, a timing problem arises with this simple implementation of the OR motif. This is because once the OR-condition is restored, neuron $I$ should be silenced immediately, along with all PSPs it is still causing in principal neurons. Clearly, the latter cannot be achieved by inhibiting neuron $I$ because a spike of neuron $I$ is an irreversible event, and PSPs, once elicited, have a fixed time course which cannot be stopped. 

A refined implementation of the OR motif therefore contains in addition an auxiliary neuron $II$ to mitigate this timing problem. Neuron $II$ is activated precisely when PSPs of neuron $I$ should be stopped: whenever some principal neuron just fired in response to neuron $I$, but the PSP of neuron $I$ is still active in other neurons. Neuron $II$ then immediately emits a spike which inhibits the principal neurons (with negative synaptic weight $-w_{OR}$), thereby canceling the effect of the prolonged PSPs of neuron~$I$.

Analogous to the WTA circuit, the described implementation of the OR circuit motif aims to approximate the requirements of Theorem~1 for modularity. To see this, consider the energy function
\begin{align}
U_{\textnormal{OR}[\mathcal{K}]} (\mathbf{x}) = \begin{cases}
	0\;, & \textnormal{if }\sum_{k \in \mathcal{K}} x_k \ge 1\;,\\
	w_{\textnormal{OR}}\;, & \textnormal{if }\sum_{k \in \mathcal{K}} x_k = 0 \;.
\end{cases}
\end{align}
Using Theorem~1, according to condition given by equation \prettyref{eq:deltaudeltae} this energy function can be implemented in a modular fashion by an auxiliary circuit making membrane potential contributions to each principal neuron in~$\mathcal{K}$,
\begin{align}
 \Delta u_{k,\textnormal{OR}[\mathcal{K}]}(t) &=  U_{\textnormal{OR}[\mathcal{K}]} (x_k=0, \mathbf{x}_{\setminus k}) - U_{\textnormal{OR}[\mathcal{K}]} (x_k=1, \mathbf{x}_{\setminus k})\\
&= 
\begin{cases}
0\;, &\sum_{l \in \mathcal{K}\setminus k} x_l(t) \ge 1\;,\\
w_{\textnormal{OR}}\;, &\sum_{l \in \mathcal{K}\setminus k} x_l(t) = 0\;.
\end{cases}
\end{align}

The OR circuit approximates this behavior as described above through the combination of two auxiliary neurons. The nature of the approximation is three-fold. First, when all principal neurons in an OR circuit have just turned off (and thus the constraint is not met anymore), the additional bias $w_{OR}$ should ideally be communicated instantly to all neurons. However, the first auxiliary neuron fires in general with some small delay, and therefore the additional bias $w_{OR}$ is signaled to the principal neurons slightly later than ideally required. Second, when 
a principal neuron eventually fires in response to the first auxiliary neuron, there is a delay until the second auxiliary neuron turns on to cancel the bias $w_{OR}$ that is still present due to lingering PSPs from the first auxiliary neuron. Third, there is an ``undershoot'' effect when the excitatory PSP of the first auxiliary neuron has already vanished, but the inhibitory PSP of the second auxiliary neuron is still present. To minimize the error due to this effect, the overall biases of all principal neurons in an OR circuit should be kept high, in order to keep the typical delay between the activity onset of the first and the second auxiliary neuron as short as possible.

\section{Details to Principle 3: Benefits of the asymmetry of spike-based stochastic search}\label{sec:Principle3}

Principles 1 and 2 pave the way towards massively parallel realizations of stochastic search in networks of spiking neurons. A first application of these principles has provided compelling results in simulations, as demonstrated in Fig.~\ref{fig:2} and Fig.~\ref{fig:4}. A key theoretical question which then arises is to what extent different components of the system contribute to the observed performance. There are various aspects that can be examined in this context, such as the asynchronicity of message transfer, stochasticity, and the asymmetry of spike-based communication (a spike marks the onset of a fixed-length \emph{on} period, whereas \emph{off} periods vary randomly - hence \emph{on} and \emph{off} states are handled fundamentally different by a spiking network). We focus our analysis here on the role of the asymmetry of spike-based signaling, because its implications are arguably least well understood.

\subsection{Asymmetric vs. symmetric dynamics}

In order to isolate the effect of asymmetric signaling we consider an artificial non-spike-based ``symmetrized'' system in which \emph{on} and \emph{off} transitions of units are not mediated in an asymmetric fashion via spikes of fixed length, but rather in a symmetric manner. Specifically, we aim to morph neural spiking dynamics into the dynamics of Gibbs sampling \cite{Bishop:06}, one of the standard methods in statistics and machine learning for sampling from complex probability distributions. By theoretically analyzing and comparing the behavior of the two systems one can then reason about the specific role of asymmetric signaling.

A canonical way of symmetrizing the dynamics of a given spiking network with noise is to make sure that all other components and aspects of the system remain unchanged (event-based asynchronous signaling, stochasticity, synaptic weights and biases, definition of membrane potential $u_k$ given the current \emph{on}/\emph{off} states of other neurons) and modify only the way the system handles transitions between \emph{on} and \emph{off} states. Importantly, to facilitate a comparison between asymmetric vs. symmetric dynamics, such modification should not alter the stationary distribution and energy function of the system. 

For a stochastic spiking neuron embedded in some network, transitions occur from \emph{off} to \emph{on} states  according to
\begin{align}
 \rho_{\textnormal{on}}(u_k) = \frac{1}{\tau} \exp(u_k)\;\;,
\end{align}
whereas transitions from \emph{on} to \emph{off} occur deterministically after a period of $\tau$ time units has passed.
Clearly, in a symmetric system transitions must occur stochastically in both directions (they cannot be both deterministic), with transition rates $\rho^{sym}_{\textnormal{on}}(u_k)$ and $\rho^{sym}_{\textnormal{off}}(u_k)$. Concrete symmetric expressions for $\rho^{sym}_{\textnormal{on}}(u_k)$ and $\rho^{sym}_{\textnormal{off}}(u_k)$ are obtained by using a continuous-time variant of Gibbs sampling \cite{Bishop:06}:
\begin{align} 
 \rho^{\textnormal{sym}}_{\textnormal{on}}(u_k) &= \rho_0 \cdot \frac{p(x_k=1,\bx_{\setminus k})}{p(x_k=1,\bx_{\setminus k}) + p(x_k=0,\bx_{\setminus k})}\\
&= \rho_0 \cdot \frac{1}{1+p(x_k=0,\bx_{\setminus k})/p(x_k=1,\bx_{\setminus k})}\\
&= \rho_0 \cdot \left(1+\exp\left(E(x_k=1,\bx_{\setminus k})-E(x_k=0,\bx_{\setminus k})\right)\right)^{-1}\\
&= \rho_0 \cdot \sigma\left(E(x_k=0,\bx_{\setminus k})-E(x_k=1,\bx_{\setminus k})\right)\\
&= \rho_0\cdot\sigma(u_k)\;\;,\\
\rho^{\textnormal{sym}}_{\textnormal{off}}(u_k) &= \rho_0 \cdot \frac{p(x_k=0,\bx_{\setminus k})}{p(x_k=1,\bx_{\setminus k}) + p(x_k=0,\bx_{\setminus k})}\\
&= \rho_0 \cdot \left(1+\exp\left(E(x_k=0,\bx_{\setminus k})-E(x_k=1,\bx_{\setminus k})\right)\right)^{-1}\\
&= \rho_0\cdot\sigma(-u_k)\;\;,
\end{align}
where $\sigma(u)=(1+\exp(-u))^{-1}$ denotes the standard sigmoid function and $\rho_0$ is an arbitrary constant controlling the global speed of sampling. Such a continuous-time variant of Gibbs sampling has been proposed in the literature, for example, in the context of sampling from second-order Boltzmann machines \cite{yamanaka1997continuous}.

\subsection{Asymmetry facilitates transitions across large energy barriers}

A somewhat unexpected difference which emerges from the comparative analysis between  asymmetric and symmetric dynamics is that transitions across large energy barriers are much more likely and frequently to occur with asymmetric (spike-based) signaling. To see this, define the mean \emph{on}-transition time $m_{\textnormal{on}}(u)$ as the average time from the last \emph{on}$\rightarrow$ \emph{off} transition until the next \emph{off}$\rightarrow$\emph{on} transition, at a given membrane potential $u$. The mean \emph{off}-transition time is defined in an analogous manner. In the stochastic spiking network these are given by,
\begin{align}
 m_{\textnormal{on}}(u) &= \frac{1}{\rho_{\textnormal{on}}(u)} = \tau\cdot \exp(-u)\;\;,\\
m_{\textnormal{off}}(u)&=\tau\;\;.
\end{align}

In the symmetric system, on the other hand, mean transition times in continuous time are given by,
\begin{align}
 m^{\textnormal{sym}}_{\textnormal{on}}(u) &= \frac{1}{\rho^{\textnormal{sym}}_{\textnormal{on}}(u)} =  \frac{1}{\rho_0}\cdot (1+ \exp(-u))\;\;,\\
 m^{\textnormal{sym}}_{\textnormal{off}}(u) &= \frac{1}{\rho^{\textnormal{sym}}_{\textnormal{off}}(u)} =  \frac{1}{\rho_0}\cdot (1+ \exp(u))\;\;.\label{seq:msymoff2}
\end{align}

\noindent Notably, one can identify a single translation factor $F(u)$ between the two systems,
\begin{align}
\frac{m^{\textnormal{sym}}_{\textnormal{on}}(u)}{m_{\textnormal{on}}(u)} &= \frac{\rho_0^{-1} \cdot (1+ \exp(-u))}{\tau \cdot \exp(-u)}\\
&= \frac{\rho_0^{-1} \cdot (1 + \exp(u))}{\tau } = F(u)\label{seq:Fon}\\
\frac{m^{\textnormal{sym}}_{\textnormal{off}}(u)}{m_{\textnormal{off}}(u)} &= \frac{\rho_0^{-1} \cdot (1+\exp(u))}{\tau} = F(u) \label{seq:Foff}
\end{align}
which is given by,
\begin{align}
F(u) = \underbrace{\frac{1}{\tau \rho_0}}_{\textnormal{const.}} \cdot \left (1 + \exp(u)\right)\;\;.\label{seq:Fu2}
\end{align}

Note that $F(u)$ is strictly positive and increases monotonically with increasing membrane potential~$u$. Furthermore, note that large values $F(u)$ signify that the spike-based dynamics is fast in comparison with the symmetric dynamics. Hence, for larger $u$ transition times are considerably shortened in the spike-based system. 
In other words, the asymmetric dynamics of spiking neurons increases specifically the \emph{on}- and \emph{off}-transition rates of those neurons with high membrane potentials~$u$ (i.e.~neurons with strong input and/or high biases). This makes sense since \emph{off} transitions in the symmetric case can be arbitrarily slowed down for large $u$, see equation \eqref{seq:msymoff2}, whereas the spike-based system will necessarily fall back to an \emph{off} state on a regular basis regardless of $u$. 

According to equation \prettyref{eq:deltaudeltae}, high membrane potentials $u$ reflect large energy barriers in the energy landscape. Therefore, the increase of transition rates for large $u$ in the spike-based system (due to deterministic \emph{on}$\rightarrow$\emph{off} transitions) means that large energy barriers are crossed more frequently than in the symmetric system. In particular, \emph{on}$\rightarrow$\emph{off} transitions to high-energy states become considerably more likely due to equation \eqref{seq:Foff}. Nevertheless, it should be stressed that on average the spike-based system does not spend more \emph{time} in high-energy states (both systems sample from the same $p(\mathbf{x})$), because according to equation \eqref{seq:Fon} also the transition back to the corresponding \emph{on} state (i.e.~the lower energy state) happens at an increased rate for large $u$. The critical observation is that the return to the identical previous state \emph{can} be intercepted: While the neuron is \emph{off}, other neurons are given the brief opportunity to spike before the previous state is restored, and may thereby, e.g., escape from a previously inhibited state. This is particularly obvious in the context of WTA circuits, where such brief periods of \emph{off}-time of the current winner allow other neurons to take over.  Altogether, as we demonstrated in Fig.~\ref{fig:2}, it is observed that this enhanced utilization of exploratory moves leads to improved search for low energy states in the asymmetric spike-based system, by facilitating fast escape routes from deep local minima which are not available to such extent in a symmetric system.

\subsection{Asymmetry facilitates goal-directed transitions}

Equation \eqref{seq:Fu2} implies that spike-based transition frequency is enhanced in proportion to $u$. It was already noted above that this encourages exploratory \emph{on}$\rightarrow$\emph{off} transitions which may facilitate the escape from local minima. But clearly also \emph{off}$\rightarrow$\emph{on} transitions are affected by equation \eqref{seq:Fu2}. In particular, consider a situation where a group of neurons in the \emph{off} state is competing for emitting the next spike (e.g. in a WTA circuit). Those neurons with the highest membrane potentials are particularly eager to fire. Suppose, for example, that there are two neurons with $u_a=3$ and $u_b=5$, and all other neurons have considerably lower $u$. According to equation \eqref{seq:Fon}, transition rates in the spike-based system are increased to a greater extent in neurons with higher $u$: In the symmetrized non-spiking system, transition rates scale with $\sigma(u)$ and are therefore approximately equal for the two neurons $a$ and $b$ (due to saturation of the sigmoid function). In the spike-based system, however, instantaneous transition rates scale with $\exp(u)$ and thus the competition will be much easier to win by the neuron which is most eager to fire (i.e.~neuron $b$ in the example). Clearly, this makes a substantial difference in the dynamics and performance of the stochastic search, especially since $u_k$ reflects the drop in energy that can be gained by turning on some neuron $k$. In particular, it means that a spike-based system is not only more exploratory in the ``up-hill'' direction (\emph{on}$\rightarrow$\emph{off} transitions towards higher energy levels), but also more goal-directed in the ``down-hill'' direction. 

Obviously, the enhanced agility with respect to some transitions must come at a price. Indeed, those transitions which bring about only small changes in the energy landscape (transitions with small) are considerably disadvantaged by the spike-based dynamics. In terms of convergence properties, however, this seems to be a small price to pay, since stochastic search appears in practice more frequently impeded by the presence of large energy barriers. \footnote{In principle, one sees from equation \eqref{seq:Fu2} that on the other hand transitions with negative $u$ are disadvantaged by the spike-based dynamics. 
In the context of this paper negative $u$ only occur in neurons which are currently inhibited in a WTA circuit.}

\subsection{Comparison of solving TSP by a spiking network and by a Boltzmann machine}\label{subsec:TSP}

Boltzmann machines are artificial (non-spiking) neural networks with stochastic binary units which have biases $b_k$ and where connections between units are required to be symmetric, i.e. $w_{kl}=w_{lk}$, \cite{hinton1983optimal,HintonSeAc84}. In order to facilitate a fair comparison between spiking network and Boltzmann machines, we adapted the previously described TSP network architecture to include only symmetric weights between neurons. This can be done by replacing di-synaptic inhibitory links mediated through WTA auxiliary neurons by direct inhibitory connections among principal neurons (in violation of Dale's law). The activation of a principal neuron in a WTA circuit will therefore automatically and directly inhibit all other principal neurons in the same WTA. We note that this adapted architecture has virtually the same energy function as the original implementation with auxiliary inhibitory neurons. 

The above described simplification allowed us to perform a fair comparison between spiking network~(SN) and Boltzmann machine~(BM), since the adapted architecture can be simulated with identical parameters (and leads to an identical stationary distribution $p(\bx)$) on both systems.
The comparison of the number of state changes between SN and BM implementations was done based on 100 runs, each of which was simulated for 100.000 state changes. Both systems were initialized at the beginning of each run in an all-silent state (i.e.~all $z_k=0$). In each run and after every state change we evaluated the current network state, and checked how many problem variables were properly defined. Combining all runs in each case, we calculated how often transitions occurred in each sampler to states with different numbers $N_{undef} \in \{0,\dots, N'\}$ of undefined problem variables.  Based on this information we constructed corresponding histograms for SN and BM. To highlight the differences between the two implementations, we calculated the ratios between the normalized histogram values for SN and BM (Fig.~\ref{fig:2}C). For the convergence speed comparison, in each run we calculated after each state change the cumulative minimum and mean performance during the whole time leading up to that state change. This was first done for each of the 100 network runs individually. The results for each number of state changes were then averaged over all runs.


\subsection{Further details to the observed spike-based energy jumps in Fig.~\ref{fig:3}C}\label{subsec:observed-jumps}


The histograms in Fig.~\ref{fig:3}C display a striking symmetricity of state transitions with positive and negative energy differences. For the Boltzmann machine this is an expected consequence of detailed balance in Gibbs sampling \cite{brooks2010handbook}, which ensures that on average  transitions between any two states are observed equally often in either direction.
For the spiking network, however, the observed symmetricity cannot be predicted from theoretical considerations alone.
This is because detailed balanced does not hold in a strict sense in the spiking network when off-transitions follow a deterministic decay law \cite{BuesingETAL:11}. Nevertheless, the spiking network dynamics can be easily morphed into a system that obeys detailed balance: When off-transitions are made stochastic with constant rate $\tau^{-1}$ (such that on average neurons are on for $\tau$ time units before turning off as in the spiking network), the dynamics of the system can be described in terms of a simple Master equation,
\begin{align}
 \frac{d p_t(\mathbf{x})}{dt} &= \tau^{-1} \sum_{k} \left[p_t(\neg x_k, \mathbf{x}_{\setminus k}) \cdot \exp(x_k u_k) - p_t(\mathbf{x}) \cdot \exp((1-x_k) u_k) \right]\\
&= \tau^{-1} p_t(\mathbf{x}) \sum_{k} \exp(x_k u_k) \left[\frac{p_t(\neg x_k| \mathbf{x}_{\setminus k})}{p_t(x_k| \mathbf{x}_{\setminus k})} - \exp((1-2x_k) u_k) \right]\\
&= \tau^{-1} p_t(\mathbf{x}) \sum_{k} \exp(x_k u_k) \left[\left(\frac{p_t(\neg x_k| \mathbf{x}_{\setminus k})}{p_t(x_k| \mathbf{x}_{\setminus k})}\right)^{1-2x_k} - \exp(u_k)^{1-2x_k}  \right]
\end{align}
where $p_t(\mathbf{x})$ denotes the probability that the system is in state $\mathbf{x}$ at time $t$, and $p_t(\neg x_k, \mathbf{x}_{\setminus k})$ denotes the probability of state $\mathbf{x}$ with negated element $k$. Detailed balance requires that all summands (i.e. the probability flows for all neurons $k$) are zero at equilibrium, for all possible states $\mathbf{x}$. It is easy to verify that this holds if the NCC given by equation \prettyref{eq:NCC} is fulfilled. Hence, although the spiking network is not strictly in detailed balance, a slightly modified dynamics of the spiking network is. This provides a potential explanation for the symmetricity of the top histogram in Fig.~\ref{fig:3}C. 

Another peculiarity of the top histogram in Fig.~\ref{fig:3}C (energy jumps in the spiking network) is its highly non-Gaussian shape (the distribution of positive jumps is even bi-modal). It should be stressed that the particular shape of the histogram, including the dip around $15$, does not reveal general properties of the spiking network dynamics, but reflects the statistics of the problem instance. Concretely, the fact that positive jumps are bi-modally distributed arises from the architecture of the TSP network (Fig.~\ref{fig:2}A): Each neuron has at most two neighbors that provide positive input. When only one neighbor is active, the membrane potential is typically between $10$ and $14$, and when both are active, the membrane potential is typically between $20$ and $28$.  The membrane potential $u$ of a neuron encodes the energy difference of an off-transition, and $-u$ represents the energy difference of an on-transition. Hence, the observed bi-modally distributed energy jumps in the spiking network are a direct consequence of the specific problem architecture. In the Boltzmann machine this effect is not visible since transitions in the relevant range of energy differences are suppressed.


\section{Details to Principle 4: Internal temperature control}

In order to realize an internal temperature control mechanism for regulating the temperature $T$ of the network energy function according to $E_T(\bx) = E(\bx) / T$ in an autonomous fashion, at least three functional components are required (Fig.~\ref{fig:1}D): 1. Internally generated feedback signals from circuit motifs reporting on the quality and performance of the current tentative network solution. 2. A \emph{temperature control} unit which integrates feedback signals and decides on an appropriate temperature $T$. 3. An implementation of the requested temperature change in each circuit motif. 

Both circuits motifs, WTA and OR, can be equipped quite easily with the ability to generate internal feedback signals. The WTA condition in a WTA circuit is met if exactly one principal neuron is active in the circuit. If the WTA circuit has strong inhibition then network states with two or more active neurons are very unlikely and can be ignored in practice. Thus, a WTA feedback signal can be generated by simply checking whether \emph{any} principal neuron is active. Concretely, the output of the auxiliary WTA neuron can be used as a feedback signal, since the auxiliary neuron is only activated when one of the principal neurons has fired. Alternatively, for the same reasons the WTA feedback signal may be constructed from the summed outputs of all principal neurons in a WTA circuit. 

For the OR circuit motif, the OR condition is met if at least one principal neuron is active. The most straightforward way of implementing an OR feedback signal is therefore to add a feedback neuron which fires as long as one of the principal neurons is active, and remains silent otherwise. This can be achieved in a straightforward manner by a feedback neuron with low bias and excitatory connections from all involved principal neurons. In simulations a slightly different implementation has proved even more effective, which can be used when all principal neurons involved in the OR circuit are also part of some WTA circuit. Then, a negative feedback signal (one which is active when the OR condition is violated) can be generated by adding an auxiliary neuron with low bias which receives connections from all other neurons in the WTA circuits of the involved principal neurons. The rationale behind this implementation of an OR feedback signal is as follows: Whenever some principal neuron $k$, which is involved in the OR circuit and in addition in some WTA circuit, is not active, most of the time some other neuron in the WTA circuit of neuron $k$ must be active. Hence, whenever the OR constraint is violated and all $K$ principal neurons involved in the OR circuit are inactive, the feedback neuron will see that in each of the involved WTA circuits some other neuron is active. A more detailed description of how this was implemented as part of the temperature control mechanism for 3-SAT problems is given in Section ''Details to 3-SAT application``.

Regarding the temperature control unit, one can think of various smart strategies to integrate feedback signals in order to decide on a new temperature. In the simplest case, a temperature control unit has two temperatures to choose from: one for exploration (high temperature), and for stabilization of good solutions (low temperature). A straightforward way of selecting a temperature is to remain at a moderate to high temperature (exploration) by default, but switch temporarily to low temperature (stabilization) whenever the number of positive feedback signals exceeds some threshold, indicating that almost all (or all) constraints in the circuit are currently fulfilled.

Concretely, such internal temperature control unit can be implemented via a temperature control neuron with a low bias and connection strengths from feedback neurons in each circuit in such a manner that the neuron's firing probability reaches non-negligible values only when all (or almost all) feedback signals are active. When circuits send either positive or negative feedback signals, the connection strengths from negative feedback neurons should be negative and can be chosen in such a manner that non-negligible firing rates are achieved only if all positive feedback but none of the negative feedback signals are active. Whenever such temperature control neuron is active it indicates that the circuit should be switched to the low temperature (stabilization) regime.

\bibliography{CSP.bib}